\documentclass{article}
\usepackage{amsmath}
\usepackage{graphicx} % Required for inserting images
\usepackage{subcaption}
\usepackage{float}

\providecommand{\keywords}[1]
{
  \small	
  \textbf{\textit{Keywords---}} #1
}

\usepackage{authblk}

\title{Graph Neural Networks in Histopathology: Emerging Trends and Future Directions}
\author[1]{Siemen Brussee}
\author[1]{Giorgio Buzzanca}
\author[1]{Anne M.R. Schrader M.D.}
\author[1,2]{Jesper Kers M.D. }

\affil[1]{Leiden University Medical Center, The Netherlands}
\affil[2]{Amsterdam University Medical Center, The Netherlands}

\date{20-06-2024}

\begin{document}

\maketitle

\begin{abstract}
Histopathological analysis of Whole Slide Images (WSIs) has seen a surge in the utilization of deep learning methods, particularly Convolutional Neural Networks (CNNs). However, CNNs often fall short in capturing the intricate spatial dependencies inherent in WSIs. Graph Neural Networks (GNNs) present a promising alternative, adept at directly modeling pairwise interactions and effectively discerning the topological tissue and cellular structures within WSIs. Recognizing the pressing need for deep learning techniques that harness the topological structure of WSIs, the application of GNNs in histopathology has experienced rapid growth. In this comprehensive review, we survey GNNs in histopathology, discuss their applications, and explore emerging trends that pave the way for future advancements in the field. We begin by elucidating the fundamentals of GNNs and their potential applications in histopathology. Leveraging quantitative literature analysis, we identify four emerging trends: \textit{Hierarchical GNNs}, \textit{Adaptive Graph Structure Learning}, \textit{Multimodal GNNs}, and \textit{Higher-order GNNs}. Through an in-depth exploration of these trends, we offer insights into the evolving landscape of GNNs in histopathological analysis. Based on our findings, we propose future directions to propel the field forward. Our analysis serves to guide researchers and practitioners towards innovative approaches and methodologies, fostering advancements in histopathological analysis through the lens of graph neural networks.
\end{abstract}
\section{Introduction}
%Evolution of computational histopathology
\par{Histopathology analysis is an important diagnostic tool and examination tool that can be used for disease diagnosis, estimating disease prognosis, and selecting for or monitoring of therapeutic strategies. Since the digitization of whole slide images (WSIs) in the early 2000s, the computational analysis of histopathology images has become an increasingly important part of histopathology. Starting with image analysis algorithms, the field transitioned to a deep learning approach following the rise of convolutional neural networks in the 2010s, largely due to the availability of large datasets (e.g., ImageNet \cite{deng2009imagenet}) and deeper convolutional architectures (e.g., AlexNet \cite{krizhevsky2012imagenet}). In the last 5 years, paradigms in the field have become more heterogeneous, with the advent of attention-based multiple instance learning \cite{ilse2018attention} \cite{sudharshan2019multiple}, vision transformers \cite{dosovitskiy2020image} \cite{wang2021transpath}, self-supervised learning \cite{chen2020simple} \cite{ciga2022self} and graph neural network \cite{scarselli2008graph} \cite{li2018graph} approaches.} \\ 
%Introduction to GNNs
\par{The emergence of Graph Neural Networks (GNNs) \cite{scarselli2008graph} has allowed effective modeling of naturally graph-structured data, such as social networks, (bio)chemical molecules \cite{schutt2018schnet} \cite{li2021structure}, geospatial data \cite{cui2019traffic} \cite{zhu2020understanding}, and tabular data which can be effectively modeled as a graph, such as in recommendation systems \cite{ying2018graph} and drug interactions \cite{zitnik2018modeling}. GNNs can be effectively used for problems involving pairwise interactions between entities in data. In addition, the topological inductive bias that can be encoded in the graph structure allows GNN models to learn based on the topology of the problem. We can define the graph neural network as an optimizeable transformation on all graph attributes that preserves graph symmetries by being permutation invariant \cite{sanchez2021gentle}. Fundamental for the graph neural network is the notion of \textit{message-passing} in which we use a learned transformation that exchanges feature information between entities in the graph, leading to topology-aware feature vectors. How the message-passing function is defined is dependent on the type of GNN used, of which many varieties exist (e.g., GCN \cite{kipf2016semi}, GAT \cite{velivckovic2017graph}, GIN \cite{xu2018powerful}). In 2018, GNNs were also introduced to histopathology \cite{li2018graph} and have gained tremendous popularity in the field since then.} \\ 
%Why this paper?
\par{While review papers on the application of GNNs in histopathology exist, they give a general overview \cite{ahmedt2022survey} or focus on the clinical applications of GNNs in histopathology \cite{meng2023clinical}. Instead, we focus on identifying and quantifying emerging trends in the application of GNNs in histopathology and use these to provide future directions in the field.} \\ 
\par
\begin{figure}[H]
    \centering
    \includegraphics[scale=0.4]{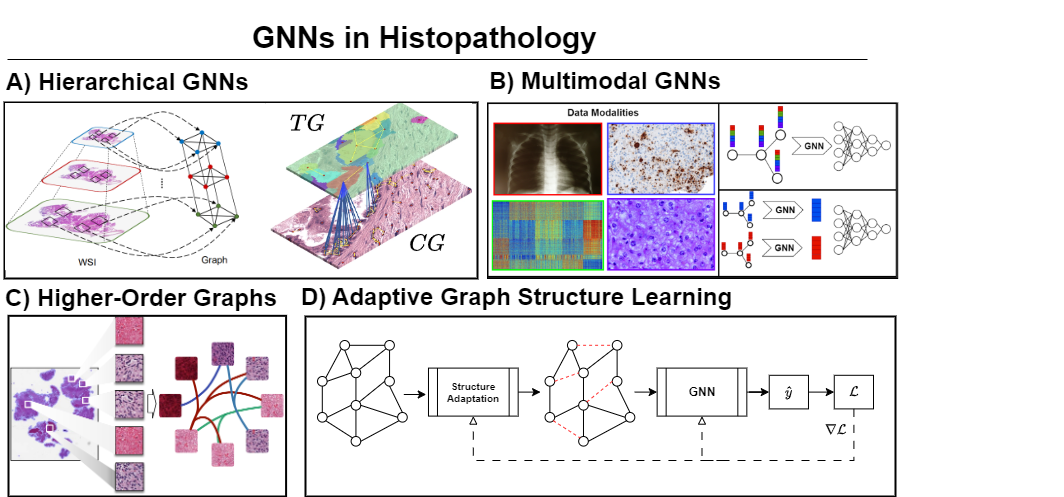}
    \caption{Overview of the four emerging subtopics of GNNs in Histopathology, covered in this review: Hierarchical GNNs, Multimodal GNNS, Higher-order Graphs, and Adaptive Graph Structure Learning. \cite{mirabadi2024grasp} \cite{pati2020hact} \cite{chen2020pathomic} \cite{di2022big} \cite{zhu2021deep}}
    \label{fig:graphabstract}
\end{figure}

{Our review is organized into three main sections: First, we introduce GNNs, and their applications in histopathology. Secondly, we identify emerging trends in the application of GNNs in histopathology, from which we select some emerging paradigms which we discuss in more depth (Figure \ref{fig:graphabstract}). Thirdly, based on our findings, we provide future directions for the field.}

\section{Graph Neural Networks in Histopathology}
\subsection{Graph Neural Networks}
\par{A graph $G$ is defined as a set of nodes $N$ connected by edges $E$: $G=(V,E)$. The set of edges is defined as a tuple of nodes: $E=\{ (x,y) | x,y \in V \}$. The connectivity of the nodes in a graph is captured in the adjacency matrix $A^{n \times n}$, where $n$ is the number of nodes in $G$.  Each entry $a_{ij} \in A$ denotes the existence of an edge $e_{ij} \in E$ as follows:
\begin{equation}
    a_{ij} = \begin{cases}
        1, & \text{if } e_{ij} \in E \\
        0, & \text{if } e_{ij} \notin E \\
    \end{cases}
\end{equation}
Alternatively, the values of $a_{ij}$ can denote edge weights ranging from 0 to 1, which represents the connectivity strength between nodes $i$ and $j$.
Given an undirected graph $G=(V,E)$, we can define the $k$-neighborhood of any node $v \in V$, noted as $N_k(v)$ recursively as follows:
\begin{align}
N_0(v) &= \{v\}, \\
N_1(v) &= \{u \mid (v, u) \in E \text{ or } (u, v) \in E\}, \\
N_k(v) &= \{u \mid \exists w \in N_{k-1}(v) \text{ such that } (w, u) \in E \text{ or } (u, w) \in E\}.
\end{align}

GNNs aggregate feature information from the $k$-neighborhood of each node, where $k$ directly corresponds to the number of GNN layers used. This aggregated information is used to update the node feature representation, $h$, in each GNN layer. Mathematically, the node representation update is defined as follows:
\begin{equation} \label{eq:abstract_gnn}
\begin{split}
    h_u^{k+1} & = \text{UPDATE}^{(k)} (h_u^{(k)}, \text{AGGREGATE}^{(k)}(\{ h_v^{(k)}, \forall v \in N_k(u)\})) \\
    & = \text{UPDATE}^{(k)} (h_u^{(k)}, m^{(k)}_{N(u)})
\end{split}
\end{equation}
where UPDATE and AGGREGATION  denote the functions that update node representation $h_u$ and aggregate the hidden representations from $u$'s neighborhood $N_k(u)$, respectively. How the UPDATE and AGGREGATION functions are exactly defined is dependent on the message passing scheme used and are usually parameterized by two learnable weight matrices. However, all message passing schemes employ a permutation-invariant AGGREGATION function (e.g., sum, mean). We can generally distinguish two types of message-passing schemes: \textit{Spectral} message-passing, based on the spectral graph properties (e.g., eigenvalues) calculated using the graph Fourier transform, and \textit{Spatial} message-passing, which are directly applied on the connectivity structure present in the input graphs. In this review, we mainly focus on spatial message-passing methods as these are applied in the vast majority of histopathology applications using GNNs. We first denote a tuple $(G,A,X)$, where $G$ denotes the input graph, $A$ the associated adjacency matrix and $X$ the input node feature matrix. To make the graph representation less sensitive to node degrees, we can normalize the adjacency matrix into a normalized adjacency matrix $\tilde{A}$, as follows:
\begin{equation}
\tilde{A} = D^{-1/2} A D^{-1/2}
\end{equation}, where $D$ denotes the degree matrix (diagonal matrix where $D_{ii}$
  is the degree of node
$i$) of the graph. To utilize spectral information in the graph structure, we can use the Laplacian matrix of the graph, defined as: $L = D - A$. During message passing, we transform feature matrix $X$ into hidden feature representation matrix $H$, typically using a learned weight matrix $W$ and a nonlinear activation function $\sigma$. 

One of the most widely adopted and earliest spatial GNN schemes is the Graph Convolutional Network (GCN). The message passing function uses a normalized adjacency matrix to update the hidden representations of nodes based on the node neighborhood. To acquire the hidden representation matrix $H$, the message passing function in GCN layer $l$ is defined as follows: 
\begin{equation}
    H^{l+1} = \sigma ( \tilde{D}^{-\frac{1}{2}} \tilde{A} \tilde{D}^{-\frac{1}{2}} H^{l} W^l)
\end{equation}
in which $\tilde{D}$ denotes the degree matrix of $G$ and $\tilde{A}$ represents the adjacency matrix with added self-loops for each node \cite{kipf2016semi}.

The spatial Graph Attention Network (GAT) \cite{velivckovic2017graph} extends the GCN scheme by adding attention weights to each edge of the graph. This essentially allows models to learn the importance of nodes during message passing. For each edge $e_{vu}$ connecting nodes $v$ and $u$, we first calculate an attention score:
\begin{equation}
    e_{vu} = \sigma \left( \vec{a}^T \left[ W h_v^{(l)} \| W h_u^{(l)} \right] \right)
\end{equation}
$\|$ denotes concatenation and $\vec{a}$ is a trainable shared parameter vector. Using this score, we can calculate the corresponding edge attention weight as follows: 
\begin{equation}
    \alpha_{vu} = \frac{\exp(e_{vu})}{\sum_{u' \in N(v)} \exp(e_{vu'})}
\end{equation}
 We then update the hidden node representation of $h_v^{(l)} \in H^{l}$ as follows:
\begin{equation}
h_v^{(l+1)} = \sigma \left( \sum_{u \in N(v)} \alpha_{vu} \cdot W^{l} h_u^{(l)} \right)
\end{equation}

the spatial GraphSAGE method \cite{hamilton2017inductive} provides a scalable and flexible framework to decide how neighboring nodes should be aggregated. It differs from other message-passing schemes in that it samples $S$ neighbors in the neighborhood of each node, instead of using all neighbors. Given hidden node representation  $h_v^{(l)} \in H^{l}$, we can define its message-passing scheme as follows:
\begin{equation}
    h_v^{(l+1)} = \sigma\left( \mathbf{W}^{(l)} \cdot \text{AGG}^{(l)} \left( \{ h_u^{(l)} : u \in \mathcal{S}_v \} \right) \right)
\end{equation}
Where $\text{AGG}$ denotes an aggregation function at layer $l$, which can be any permutation invariant function (e.g., sum, mean).

Xu et al. introduced the Graph Isomorphism Network (GIN) \cite{xu2018powerful}, which has an expressive spatial message-passing scheme aimed to differentiate between isomorphic graph structures \footnote{%
\begin{math}
\begin{aligned}
G_1 &= (V_1, E_1) \text{ and } G_2 = (V_2, E_2) \text{ are \textbf{isomorphic} } \iff \exists f: V_1 \to V_2 \text{ such that } \\
f &\text{ is a bijection and } \forall u, v \in V_1, \{u, v\} \in E_1 \iff \{f(u), f(v)\} \in E_2.
\end{aligned}
\end{math}
}
For any hidden node representation $h_v^l \in H^l $, the message-passing is defined as follows:
\begin{equation}
h_v^{(l+1)} = \text{MLP}^{(l)} \left( (1 + \epsilon^{(l)}) \cdot h_v^{(l)} + \sum_{u \in \mathcal{N}(v)} h_u^{(l)} \right)
\end{equation}
Here, the $\text{MLP}$ denotes a multilayer perceptron which process each node's aggregated feature vector. $\epsilon^{l}$ is a learnable parameter which learns how to scale the node's own feature vector. 

The spectral ChebNet \cite{tang2019chebnet} method uses Chebyshev polynomials to approximate spectral graph convolution. First, we rescale the graph Laplacian matrix $L$ using the largest eigenvector of $L$, $\lambda_{max}$: $\hat{L} = (2L / \lambda_{max}) - I$. Given the approximation parameter $k$, we can compute the approximated Chebyshev polynomial $Z^{(k)}$ as follows:
\begin{align}\begin{aligned}\mathbf{Z}^{(1)} &= \mathbf{X}\\\mathbf{Z}^{(2)} &= \mathbf{\hat{L}} \cdot \mathbf{X}\\\mathbf{Z}^{(k)} &= 2 \cdot \mathbf{\hat{L}} \cdot
\mathbf{Z}^{(k-1)} - \mathbf{Z}^{(k-2)}\end{aligned}\end{align}
Finally, the message-passing function to update hidden representation matrix in layer $l$, $H^{l}$, is defined as follows:
\begin{equation}
\mathbf{H}^{l+1} = \sum_{k=1}^{K} \mathbf{Z}^{(k)} \cdot
\mathbf{W}^{(l)}
\end{equation}
%Mathematical definition of Graph Theory
%Message-passing schemes (Spectral vs Spatial), how are they related? 
%Prediction tasks (Node, Edge, Graph), Pooling
\par{Prediction tasks using GNNs can be categorized into node-level, edge-level, and graph-level prediction tasks. Node-level tasks, such as node classification, predict labels of target nodes based on the transformed representations after message-passing. Edge-level tasks include edge classification, where labels are predicted for edges in the graph, and link prediction. In link prediction, the aim is to predict whether links between nodes should exist based on the node features after message passing. Lastly, graph-level tasks need a global \textit{pooling} step, which aggregates information from node and / or edge level into a global representation which can be used to predict graph-level labels. Let us define a graph $G=(V,E)$ with an associated node feature matrix $X$. We can then use any permutation-invariant function to pool the node features into a global representation:
\begin{equation}
    pool(G) = \bigoplus_{v \in V} X(v)
\end{equation}
where $\bigoplus$ is any permutation invariant function (e.g., sum).
}

\subsection{GNNs in Histopathology}
Graphs have been used in digital pathology since the 1990s \cite{sharma2015review} and have later been combined with classical machine learning algorithms for diagnostic tasks \cite{bilgin2007cell}. Since then, GNNs have been gaining popularity throughout the 2010s to become the primary method for graph-based machine learning tasks. Since the first application of GNNs in histopathology, in 2018 \cite{li2018graph}, the use of GNNs in histopathology has grown rapidly, with more than 150 publications in 2024. GNNs offer several important advantages for modeling of histopathological images:
\begin{enumerate}
    \item \textit{GNNs acquire relationship-aware representations} By exchanging information between nodes in the input graph, GNNs learn context-aware representations. This is important in pathology, where meaningul biological structures often depend on the cellular or regional context \cite{santoiemma2015tumor}. It should be noted that vision transformer models do also allow learning relationship-aware representations but these relationships are calculated between arbitrary patches instead of between predefined biologically relevant entities (e.g., cells), in the case of GNNs.
    \item \textit{GNNs can learn from the topological information in the WSI} Graphs are a natural way to capture topology. In histopathology, factors like cellular density can be important in diagnosis, which can be captured in the topological information in the graph structure \cite{ali2013cell} \cite{reynolds2014cell}. 
    \item \textit{GNNs model the entire WSI at once} Due to the sheer size of whole slide images, traditional deep learning methods usually split the WSI into image patches and use these as model input. This approach introduces patching bias, as optimal resolution, size, and stride depend on the problem at hand \cite{hou2016patch}. GNNs can model the WSI as a graph, which is much smaller than the original image. This allows it to be loaded into memory, effectively capturing the global structure of the WSI \cite{adnan2020representation}.
    \item \textit{GNNs allow for hierarchical modeling} In histopathological image analysis, diagnosis often relies on information acquired from multiple spatial scales of the WSI (e.g., global patterns combined with specific cellular features). GNNs allow for modeling both these scales in a single model, either by connecting graphs on different scales or by learning the global structure through pooling operations \cite{pati2020hact} \cite{zheng2019encoding}. 
    \item \textit{GNNs allow for entity-wise interpretability} Whereas CNN-based methods usually rely on pixel-level explainability, GNNs allow for entity-wise explainability. This allows pathologists to investigate the dependence of the model prediction on certain biological entities, such as cells or substructures in the WSI \cite{sureka2020visualization}.
    \item \textit{GNNs allow for injecting task-specific inductive biases} The input graph structure can be modified based on prior information about the task at hand. This in turn allows for more specific explainability and efficient modeling of the problem \cite{hasegawa2023edge}.
    \item \textit{GNNs allow for straightforward multimodal integration} Multimodal integration often requires modeling separate modules whose information is fused together to arrive at a final prediction. In GNNs, information can be simply added to the feature vectors associated with the node, edge, or graph, which can then be jointly updated using message passing. This approach is efficient, as no additional model modules are required and allows quick injection or removal of information from different modalities \cite{ektefaie2023multimodal} \cite{li2022differentiation}.
\end{enumerate}

%What has been the role of GNNs in histopathology?
%What advantages do GNNs have in histopathology compared to other methods?
%How do GNNs perform compared to other methods in histopathology?
The application of GNNs to histopathology requires some decision making and algorithmic steps (Figure \ref{fig:gnn_workflow}). First, we preprocess the WSI (e.g., quality control, stain normalization). Now, either a cell segmentation algorithm can be applied from which a cell graph can be constructed, or one extracts patches from which a patch graph can be constructed. Using the extracted image entities, a graph can be defined using a chosen graph construction algorithm. When a graph has been defined, it can be used as input for a GNN-model. The predictions given by the GNN-model can be explained using various GNN explainability methods. We will further explore this typical workflow of GNNs in histopathology in the following sections.
\subsubsection{Defining the input graph}
\par{For GNNs to be applied to histopathology images, one first needs to define which entities nodes in the input graph will represent. The majority of GNNs applied to histopathology use one of 3 types of input graphs, as shown in Figure \ref{fig:graphtypes}: \textit{Cell Graphs}, where nodes represent cells or nuclei, segmented using a segmentation algorithm or model (e.g., HoVerNet \cite{graham2019hover}). \textit{Patch Graphs}, where nodes represent patches of the image, and lastly, \textit{Tissue Graphs}, where nodes represent larger-scale semantic entities in the graph. These tissue graph entities can be acquired from a semantic segmentation map, superpixels (usually generated using the SLIC algorithm \cite{achanta2012slic}), or clustered superpixels, which represent similar regions in the input image. Some alternate approaches also exist; notably, approaches that treat image pixels as nodes and approaches that construct a patch-based hypergraph \footnote{graph where edges can connect any number of nodes instead of the pairwise edges seen in regular graphs}. }

\begin{figure}[H]
    \centering
    \includegraphics[scale=0.45]{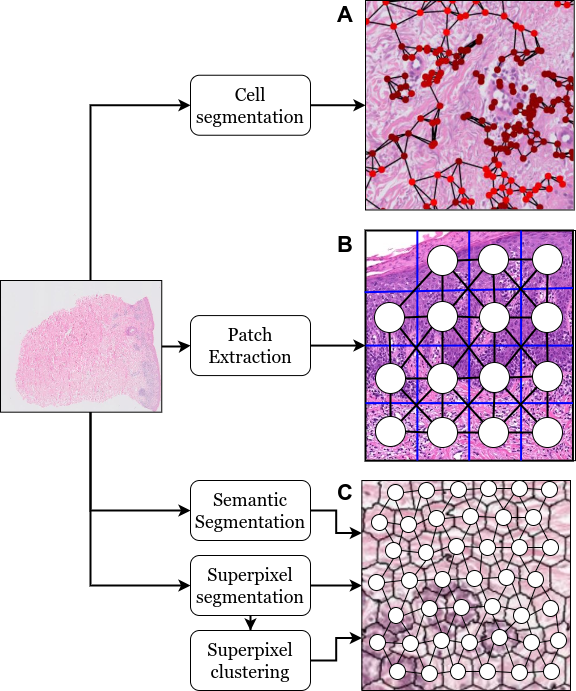}
    \caption{Most widely used graph types in GNNs for histopathology. A) Cell Graph, B) Patch Graph, C) Tissue graph (based on superpixels, clustered superpixels, or a semantic segmentation mask. The superpixel image was acquired from \cite{bejnordi2015multi}.)}
    \label{fig:graphtypes}
\end{figure}
%Cell graphs, Patch graphs, superpixel graphs
\par{Once the entities for the nodes have been established, one needs to decide how the nodes should be connected. For this, histopathology GNNs usually use one of four graph construction strategies, or combinations of these strategies. First, we can use a simple \textit{distance threshold}, where we connect all nodes having a pairwise distance (e.g., Euclidean) less than a set threshold $t$. Second, we can use the \textit{k-Nearest Neighbor} (k-NN) algorithm. Here, we set a parameter $k$, which denotes how many neighbors each node will have. Then, we connect the $k$ closest neighbors of each node to the target node. Note that for both of these approaches, we can base our notion of distance on spatial distance or distance between the node-associated feature vectors. Third, we can construct a \textit{Region Adjacency Graph} (RAG), where we connect all entities that share a border\footnote{In patch graphs, this is equivalent to using a $k=4$ k-NN without diagonal neighbors and $k=8$ k-NN with diagonal neighbors.}. Typically, this approach is used for patch- or tissue graphs, with a clear border between entities. Lastly, we can use \textit{Delaunay triangulation}. Here, we form all possible triangles between the nodes, such that the circumcircle of each triangle does not contain other nodes than the 3 nodes the triangle consists of.}

\begin{figure}[H]
    \centering
    \includegraphics[scale=0.45]{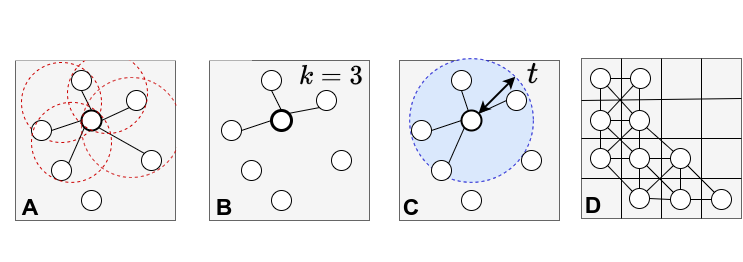}
    \caption{Most widely used graph construction techniques in GNNs for histopathology. A) Delaunay triangulation. B) K-NN with $k=3$, C) Distance threshold with threshold $t$, D) RAG with diagonal neighbors ($k=8$).}
    \label{fig:graphconstruct}
\end{figure}
%Edge connectivity (KNN, Delaunay, RAG, Distance threshold)
\subsubsection{Extracting features}
\par{To allow a GNN to use the image-based features present in whole slide images, one usually extracts features associated with the entity of the node and attaches that to the node as node features. Similarly, one can also add features to graph edges, which the GNN can use in the message-passing function. Backbones of pretrained\footnote{usually on the ImageNet dataset \cite{deng2009imagenet}} CNN (e.g., ResNet \cite{he2016deep}) or Vision Transformer \cite{dosovitskiy2020image} models are primarily used for node feature extraction, where we use an image patch corresponding with the node entity, process it using the feature extraction model, and extract the feature vectors of this image in the intermediate layers of the model as node features. Sometimes, the feature extraction model is pretrained in a supervised manner on the histopathology images for the problem itself, or fine-tuned for the prediction task at hand, which allows for more problem-specific features. More recently, self-supervised training has been applied for feature extraction, allowing for learning features that generalize better across prediction tasks \cite{tendle2021study}. \textit{Handcrafted} features, based on morphology-, texture- or intensity measurements can also be used as node features. Furthermore, (spatial) graph features (e.g., node degree) can be calculated on a node-, edge- or graph-level to more directly incorporate topological information in the model prediction.}
\subsubsection{Graph Neural Network architectures}
\par{Most message-passing schemes used in histopathology GNNs are not specific to histopathology. Popular schemes used include Graph Convolutional Networks (GCNs), Graph Attention Networks (GATs), GraphSAGE, or GINs (Graph Isomorphism Networks). Some approaches invented schemes specific for their problem \cite{gao2021gq} \cite{zhang2022ms} \cite{hou2022h} \cite{hasegawa2023edge} \cite{nakhli2023co} \cite{wang2023deep} and lately, Graph Transformers models have gained traction as a popular alternative or addition to regular message-passing. In the overall model architectures, many approaches combine message-passing layers with other neural network modules, like transformers, LSTMs, MLPs, and MIL aggregation layers. For graph-level prediction problems, global pooling layers are applied, sometimes combined with sequentially applied local pooling layers which hierarchically coarsen the graph. }
%Architectures specific to histopathology
\subsubsection{Applications}
\par{GNNs in histopathology have been applied to a wide variety of tasks. Mainly on supervised prediction tasks such as survival prediction, region-of-interest (ROI) classification, cancer grading, cancer subtyping, cell classification, and the prediction of treatment response. Some applications aim to predict data in other modalities, such as genetic mutations or (spatial) gene expression. Although most use cases are classification problems, some research has used GNNs for semantic segmentation \cite{anklin2103learning} \cite{zhang2021joint} \cite{he2023gnn} or nuclei detection \cite{bahade2023cascaded} \cite{wang2023cross}. Another interesting application is \textit{Content-Based Histopathological Image Retrieval} (CBHIR). Here, we first use GNNs to extract- and save a graph representation for a ROI in a WSI. When pathologists grade new cases, we can use these embeddings to retrieve similar ROIs, helping in the diagnostic process. Most GNN applications focus on cancer as a disease, with a few exceptions \cite{wojciechowska2021early} \cite{nair2022graph} \cite{hasegawa2023edge} \cite{gallagher2023multi} \cite{lee2023clustering} \cite{su2023prediction} \cite{acharya2024prediction}. }
%Cancer subtyping, Cancer staging, Survival prediction, CBHIR, Prediction of mutation/ST

\subsection{Explainability}
\par{One major advantage GNNs have over other model types in histopathology is interpretability. The model output can be explained on an entity level and visualized using a graph overlay. For example, one can pool nodes in a cell graph using an attention mechanism, calculate the attention scores for each node, assign a color based on the attention score per node, and then visualize the attention scores on a cellular level when overlaying the graph over the WSI. Many methods for explainability in GNNs have emerged since the inception of the GNN (e.g., GNNExplainer \cite{ying2019gnnexplainer}, GCExplainer \cite{magister2021gcexplainer}). There have also been efforts to develop explainability methods specific for histopathology GNNs \cite{jaume2020towards} \cite{yu2021towards} \cite{abdous2023ks} or to use combinations of existing GNN explainability techniques to extract a clinically interpretable model output \cite{di2023digital}. }
%Introduction to GNNs and their potential in addressing limitations of traditional methods
%Case studies demonstrating the effectiveness of GNNs in histopathological analysis

\begin{figure}[H]
    \centering
    \includegraphics[scale=0.45]{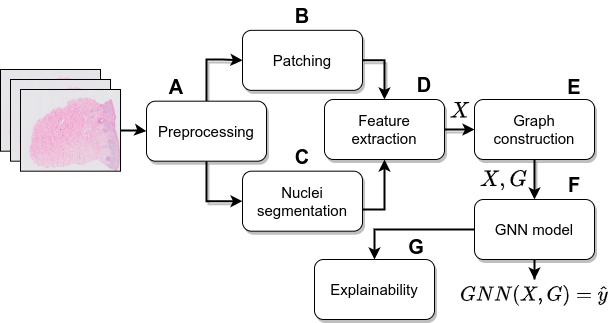}
    \caption{Overview of a typical workflow of applying GNNs to histopathology whole slide images. A) First, preprocessing steps, such as slide quality thresholds and tissue segmentation (e.g., using Otsu thresholding) are applied. B) Then, if chosen for a patch graph approach, the WSI is divided into smaller image patches. C) When a cell graph approach is used, nuclei-segmentation algorithms are applied to acquire a mask of the nuclei in the WSI. D) For each acquired entity (patch, nucleus) features are extracted, typically using a pretrained CNN-model (e.g., ResNet) to acquire a feature matrix $X$. E) Using a graph construction strategy (e.g., k-NN), entities are connected to other entities to form a cell/patch graph, $G$. F) Now, this graph, along with its associated feature matrix, can be used as input for a GNN model which applies message passing operations to learn a representation and then produces an output depending on the prediction task. G) (Graph) explainability methods can be applied to the GNN model to acquire interpretable information on the model behaviour and its predictions.}
    \label{fig:gnn_workflow}
\end{figure}
\section{Methodology}
\par{Using Google Scholar, we identified 156 papers applying GNNs to histopathology. The first of these papers is from September 2018, when the first paper applying GNNs to histopathology was published, up to March 2024. We included all papers applied on H\&E stained whole slide images or tissue microarrays (TMAs) where GNNs (i.e., message-passing) were part of the methodology. The papers were categorized based on the following properties:
\begin{itemize}
    \item Message-passing scheme
    \item Type(s) of input graph
    \item Graph construction method
    \item Feature extraction method
    \item Application(s)
    \item Tissue type(s)
    \item Hierarchy
    \item Multimodality
\end{itemize}
We quantified the frequencies in each of these properties to identify emerging trends in the literature (Figure \ref{fig:quantification}).}
\begin{figure}[H]
\hspace*{-2cm}
    \includegraphics[scale=0.3]{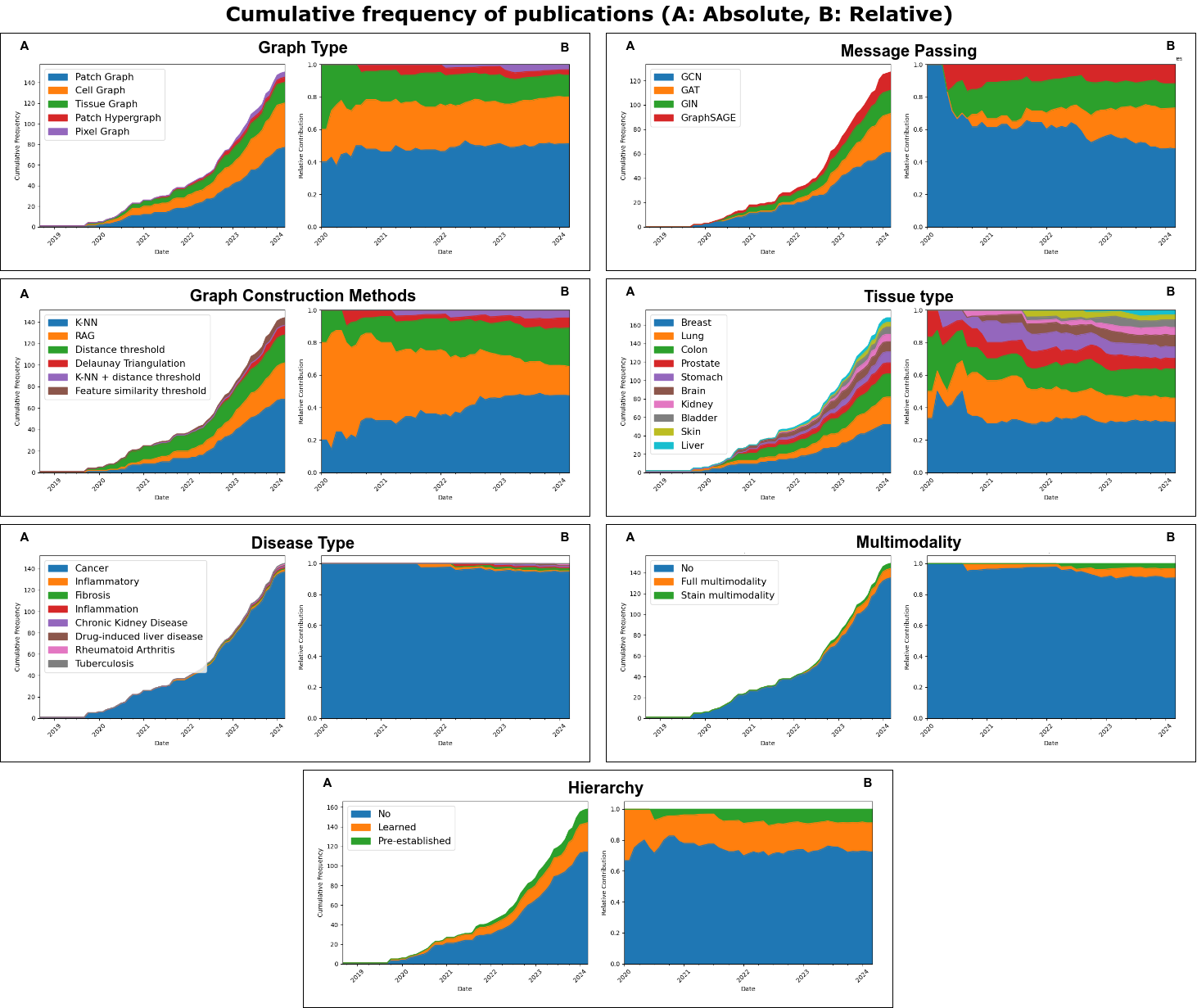}

    \caption{Cumulative frequency of publications on GNNs applied on histopathology, with different properties (e.g., Application, Graph type). For the types of message passing, graph types, graph constructors, and applications, only properties occurring in more than 4 papers were retained in the plot.}
        \label{fig:quantification}
\end{figure}

\par{From our quantification, we identified 4 upcoming trends to explore further:
\begin{enumerate}
    \item Hierarchical GNNs
    \item Adaptive Graph Structure Learning
    \item Multimodal GNNs
    \item Higher-order graphs
\end{enumerate}}
\section{Emerging Trends}
   \subsection{Hierarchical GNNs}
   \par{Diagnostic- and prognostic information present on WSIs often exists on multiple levels of coarsity. For example, the cellular microenvironment can be an important diagnostic factor but can depend on where this microenvironment is globally located in the tissue. Cellular graphs are suitable for capturing the microenvironment, but can miss the global tissue information present in the WSI. Similarly, patch- or tissue-based graphs can capture global information in the WSI, but miss the topological information of the cellular structures \cite{pati2022hierarchical}. To connect the information on different levels of coarsity, we can either apply local pooling layers which learn a hierarchical representation of the input graph in an end-to-end manner, denoted as \textit{Learned Hierarchy}, or we can define the hierarchy between graphs prior to model training, denoted as \textit{Pre-established hierarchy}. Both are illustrated in Figure \ref{fig:hierarchy}.
   \\ \\
    In a learned hierarchy, we apply \textit{local pooling} layers that can iteratively coarsen the graph structure hierarchically. Let us define our input graph with associated node features as $G_0 = (V_0, E_0, X_0)$. Assuming that we have $k$ local pooling layers in our GNN architecture, we sequentially coarsen our input graph to $G1, G_2,...,G_k$ where $G_k$ is the final pooled graph representation. Mathematically, we define a local pooling to coarsen the graph $G_i$ to $G_{i+1}$  as follows:
    \begin{equation}
            G_{i+1} = \text{pool}_i(G_i), \quad \forall i \in \{0, 1, 2, ..., k-1\}
    \end{equation}
    where $pool_i$ is defined by any permutation-invariant pooling function. Prominently used examples include DiffPool \cite{ying2018hierarchical}, SAGPool \cite{lee2019self}, and MinCutPool \cite{bianchi2020spectral}. Apart from pure local pooling, we also classify methods that learn the hierarchy using a cross-hierarchical transformer  \cite{hou2022spatial} \cite{shi2023mg} \cite{azadi2023all} layer as learned hierarchy methods. 
\\ \\
    Learned hierarchy methods learn a node assignment matrix $S^{(l)}$ which denotes the changes in the graph structure after applying the pooling operation. Often, multiple local pooling layers are applied subsequently to coarsen the graph. One sets a \textit{pooling ratio} hyperparameter, denoted as $k$, which determines how many nodes should be present after the pooling operation. For any one of these layers, $l$, the pooling operation updates the adjacency matrix of the input graph, $A$, and its corresponding node attributes $X$. The hidden representations are denoted $H$, where $X = H^{0}$. We denote the pooling operation as:
    \begin{equation}
        (A^{l+1}, H^{l+1}) = \text{POOL}(A^{l}, H^{l})
    \end{equation}
    The pooling operation is dependent on the pooling function used. DiffPool \cite{ying2018hierarchical} applies a graph neural network to learn a differentiable cluster assignment matrix which maps nodes to clusters, which are used as individual nodes after the pooling operation. DiffPool uses two GNNs: one for obtaining node embeddings, $GNN_{l,embed}$, and one for assigning the nodes to cluster nodes, $GNN_{l, pool}$. In each DiffPool layer $l$, we use the embedding GNN for extracting a feature matrix $Z$:
    \begin{equation}
        Z^{l} = GNN_{l,embed}(A^{l}, H^{l})
    \end{equation}
    Then, we calculate the assignment matrix using the pooling GNN:
    \begin{equation}
        S^{l} = softmax(GNN_{l,pool} (A^{l}, H^{l}))
    \end{equation}
    Now, we update both the hidden node representations and a new adjacency matrix:
        \begin{align}
            H^{l+1} = S^{l^T} Z^{l} \\
            A^{l+1} = S^{l^T} A^{l} S^{l}
        \end{align}

    Self-Attention Graph Pooling (SAGPool) \cite{lee2019self} uses the self-attention mechanism mechanism to learn which nodes are important and to discard unimportant ones. First, we calculate the self-attention score using a graph convolution operation: \begin{equation}
        H^{l+1} = \sigma ( \tilde{D}^{-\frac{1}{2}} \tilde{A} \tilde{D}^{-\frac{1}{2}} H^{l} W^l)
    \end{equation}
    Here, $W^l$ is a learned weight matrix which we use to calculate the attention score. For each node $v \in V$, we calculate:
    \begin{equation}
        \alpha^l_i = softmax(W^l \dot h^l_i)
    \end{equation}
    where $h_i$ is the feature embedding of $v_i$. SAGPool then ranks the nodes on their attention scores and selects the top-$k$ nodes to retain. Based on the nodes to retain, the adjacency matrix gets masked and this mask, $H_{mask}$, gets multiplied with the original adjacency matrix to coarsen the graph: $A^{l+1} = A \odot H_{mask}$.

    Lastly, MinCutPool \cite{bianchi2020spectral} uses the mincut partition objective function to decide the assignment matrix $S$. Similarly to the DiffPool method, we first generate a GNN-based node feature matrix $H^{l+1}$:
    \begin{equation}
        H^{l+1} = GNN(H^l, A^l, W^l_{GNN})
    \end{equation}
    where $W^l_{GNN}$ is the learned weight matrix of the GNN. Using the updated representation, we can use a multilayer perceptron (MLP) to calculate the node assignment matrix $S$:
    \begin{equation}
        S = MLP(H^{l+1}, W^l_{MLP})
    \end{equation}
    Where $W^l_{MLP}$ are the learned weights of the MLP. Both $W^l_{GNN}$ and $W^l_{MLP}$ are trained by minimizing two loss terms $L_c$, denoting the cut loss term, and $L_o$, denoting the orthogonality loss term. The cut loss term approximates the Mincut objective, by aiming to minimize the number of edges between clusters while maximizing the edges within clusters. The orthogonality loss term encourages orthogonal cluster assignments and similarly sized clusters. Together, these loss functions form the objective loss $L_u$:
    \begin{equation}
        L_u = L_c + L_o = -\frac{{\text{Tr}(S^\top \tilde{A}S)}}{{\text{Tr}(S^\top D\tilde{S})}} + \frac{{\text{Tr}(S^\top S - IK)}}{{\sqrt{K}}}
    \end{equation}
    Where $D$ is the degree matrix of the normalized adjacency matrix $\tilde{A}$, $I$ is the identity matrix and $K$ is the number of desired clusters. 

    The pooling operation is performed as follows:
\begin{equation}
    \begin{aligned}
        A^{l+1} &= S^T \tilde{A} S  \\
        H^{l+1} &= S^T H 
    \end{aligned}
\end{equation}

    \subsubsection{Learned hierarchy}
    As Table \ref{tab:learnedhierarchies} shows, the vast majority of GNN applications in histopathology use existing local pooling functions such as in the examples above. In this section, we give some examples of newly designed learned hierarchy methods, specifically for problems in histopathology.
    \\ \\
     \textbf{Local Pooling:} Hou et al. \cite{hou2022h} proposed \textit{Iterative Hierarchical Pooling} (IHPool), which they combined with a pre-established hierarchy. As input, the authors used a pyramidal heterogeneous patch graph, with one graph existing on 10x resolution, one on 5x resolution, and one on thumbnail resolution. Features were generated using KimiaNet. IHPool was designed to filter redundant information for the downstream prediction task while retaining this pyramidal structure when applying the pooling operation. The method achieves this by conditioning the set of nodes to be pooled on each resolution level on the pooling outcome of the lower-resolution nodes. Let $X$ be a matrix of node features, $A$ be the adjacency matrix of the input graph, $k$ be the ratio of nodes to retain after pooling and $P$ be a learnable projection layer. Now, let us denote the input graph $G=(V,E,R)$ where $R$ represents the set of different resolutions in the graph. For each resolution $r \in R$, patches on resolution $r$ are represented as nodes. The nodes are pooled hierarchically, such that nodes in higher magnification levels are subordinate to nodes in lower levels. For all nodes, a fitness score is calculated and nodes are assigned to clusters based on spatial distance and fitness difference between nodes. Specifically, for each node $n \in N$ on resolution $r$, we use a learnable projection matrix $P$ to calculate the fitness scores as follows:
    \begin{equation}
                        \phi_n^r = \tanh\left(\frac{{V_n^r \cdot P}}{{||P||}}\right)
    \end{equation}
    where $V_n^r$ is the set of nodes to be pooled, based on the hierarchical edges between resolutions. Based on the calculated node assignments, we create a new node feature matrix \( X' \). The adjacency matrix \( A' \) is updated to maintain graph connectivity based on the node assignments.
\\ \\
    Wang et al. \cite{wang2023ccf} proposed a new module for pooling information from cell graphs to use as embeddings for clusters of cells, called \textit{cell community forests}. The authors first applied DBSCAN clustering to cell embeddings where they clustered the cells based on their density. The hierarchical relationships between the cellular structures is captured by organizing the clusters into nested relationships based on their density (i.e. each dense cellular cluster is nested within a sparser, larger cluster). Cellular features pooled hierarchically up to the sparsest cluster level and then processed by a LSTM module to construct the graph embedding for downstream predictions.
    \\ \\
    Zhao et al. \cite{zhao2023mulgt} proposed an extension of the popular MinCutPool by adding an additional message-passing layer in the pooling equation. For acquiring the cluster assignment matrix $S$, where each node $s \in S$ will be a single node in the coarsened graph, the authors used the following equation: \begin{equation}
        S = H(\sigma(\hat{A} H W_{pool}))
    \end{equation}
    where $\hat{A}$ is the LaPlacian-normalized adjacency matrix $H$ denotes the hidden representation matrix of the nodes, $W_{pool}$ denotes a learnable pooling weight matrix and $\sigma$ denotes a nonlinear activation function (e.g., ReLU).
\\ \\
    \textbf{Attention-based Interaction Modeling:} Azadi et al. \cite{azadi2023all} proposed two attention-based methods for exchanging information between different levels of graph coarsity. The authors used a local graph, where nodes represent patches in the WSI, and a global graph, where nodes represent MinCutPool-based clusters of nodes in the local graph. Now, attention scores are calculated for each node in the local- and global graph. The first method the authors proposed for exchanging information between the local- and global graph was \textit{Mixed Co-Attention} (MCA), in which the information is not mixed directly, but weight sharing is applied between parallel processing of the local- and global nodes. the second method was \textit{Mixed Guided Attention}, where the idea of MCA was expanded on by directly infusing the calculated local node feature representation into the attention score calculation of the global nodes. The authors found that the mixed co-attention strategy worked optimally for their use case.
\\ \\
    \textbf{Alternative Approaches:} Ding et al. \cite{ding2023fractal} did not learn a hierarchical representation using pooling layers, instead using a \textit{FractalNet} architecture. Here, the input graph is given to separate processing paths which consist of different numbers of GNN layers, thereby representing different semantic levels in the tissue. The hierarchy between the paths is encoded using a combination of a gated bimodal unit and an MLP mixer architecture. The former calculates a weighted combination of representations, while the latter enhances communication between the path representations and strengthens connections among different path features. 
\\ \\
    Li et al. \cite{li2023hierarchical} propose a hierarchical Graph V-Net to encode hierarchy in a patch graph input. First, attention-based message-passing is used to exchange information between adjacent patches. Then, the authors used a graph coarsening operation where the node features are arranged as a 2D grid based on the spatial location of the patches. This grid is then evenly divided into submatrices and each submatrix is projected to a single feature vector using a learnable layer, which will act as a node after the coarsening operation. Notably, the Graph V-Net also uses graph upsampling layers, which add nodes until the size of the input graph has been restored, similar to what is done in UNet-architectures.

    \begin{table}[H]
        \hspace{-3cm}
        \begin{tabular}{|c|c|c|c|}
        \hline
        \textbf{Publication} & \textbf{Date} & \textbf{Application} & \textbf{Learned hierarchy method} \\
        \hline
         Zheng et al. \cite{zheng2019encoding} & 2019/10 & CBHIR & DiffPool \\
         Zhou et al. \cite{zhou2019cgc} & 2019/10 & Cancer grading & DiffPool \\
         Sureka et al. \cite{sureka2020visualization} & 2020/10 & Binary classification & DiffPool \\
         Zheng et al. \cite{zheng2020diagnostic} & 2020/12 & CBHIR & DiffPool \\
         Chen et al. \cite{chen2020pathomic} & 2020/09 & Survival prediction, Cancer grading & SAGPool \\
         Jiang et al. \cite{jiang2021weakly} & 2021/01 & Cancer grading & DiffPool \\
         Zheng et al. \cite{zheng2021histopathology} & 2021/04 & CBHIR & DiffPool \\
         Wang et al. \cite{wang2021hierarchical} & 2021/09 & Survival prediction & SAGpool \\ 
         Xiang et al. \cite{xiang2021multiple} & 2021/10 & Binary classification & DiffPool \\
         Xie et al. \cite{xie2022computational} & 2022/01 & Treatment response prediction & TopKPooling \\
         Dwivedi et al. \cite{dwivedi2022multi} & 2022/04 & Cancer grading & SAGPool \\
         Hou et al. \cite{hou2022h} & 2022/06 & Binary classification & IHPool \\
         Bai et al. \cite{bai2022scalable} & 2022/08 & Cancer subtyping & MinCutPool \\
         Zuo et al. \cite{zuo2022identify} & 2022/09 & Survival prediction & SAGPool \\
         Hou et al. \cite{hou2022spatial} & 2022/09 & Cancer subtyping & Hierarchical attention mechanism \\
         Lim et al. \cite{lim2022comparative} & 2022/10 & Survival prediction & SAGPool \\
         Wang et al. \cite{wang2023ccf} & 2023/02 & Cancer subtyping & Scattering Cell Pooling \\
         Zhao et al. \cite{zhao2023mulgt} & 2023/02 & Cancer subtyping, Cancer grading &  GCMinCut \\
         Ding et al. \cite{ding2023fractal} & 2023/02 & Cancer subtyping, Cancer grading & Fractal paths \\
         Ding et al. \cite{ding2023using} & 2023/04 & Survival prediction & SAGPool \\
         Li et al. \cite{li2023hierarchical} & 2023/09 & Node classification & Graph V-Net \\
         Syed et al. \cite{gallagher2023multi} & 2023/09 & Rheuma subtyping & SAGPool \\
         Shi et al. \cite{shi2023mg} & 2023/09 & Cancer subtyping, mutation prediction & Hierarchical attention mechanism \\
         Wu et al. \cite{wu2023transfer} & 2023/10 & Survival prediction & SAGPool \\
         Nakhli et al. \cite{nakhli2023co} & 2023/10 & Survival prediction & SAGPool\\
         Azadi et al. \cite{azadi2023all} & 2023/10 & Survival prediction & MinCutPool, Hierarchical attention mechanism \\
         Hou et al. \cite{hou2023multi} & 2023/10 & Survival prediction & Matrix multiplication \\
         Abbas et al. \cite{abbas2023multi} & 2023/12 & Cancer grading & DiffPool \\
         Xu et al. \cite{xu2023lymphoma} & 2023/12 & Cancer subtyping & DiffPool \\
         Azher et al. \cite{azher2023spatial} & 2024/01 & Cancer grading, Survival prediction & SAGPool \\
         Yang et al. \cite{yang2024prediction} & 2024/03 & Binary classification, Survival prediction & MinCutPool \\

         \hline
        \end{tabular}
        \caption{Publications applying GNNs to histopathology which used learned hierarchies}
        \label{tab:learnedhierarchies}
    \end{table}
    
    \subsubsection{Pre-established hierarchy}
    In pre-established hierarchy, we encode the hierarchy prior to model training. For example, we can construct multiple graphs at different levels of coarsity in the WSI, and connect them using assignment matrices, which denote how the nodes are connected between the hierarchical levels. During message-passing, the learned representations of the lower hierarchy level are aggregated and used as input for the corresponding nodes at the higher hierarchy level. We differentiate between approaches connecting graphs on different semantic levels (e.g., cells and tissues), and approaches connecting different magnifications of the WSI (e.g., 40x, 20x). An overview of publications using this approach is given in Table \ref{tab:establishedhierarchies}.
\\ \\
    \textbf{Semantic Hierarchies:} Pati et al. \cite{pati2020hact} were the first to introduce a pre-established hierarchy in the graph to use as input for a GNN model. They constructed a cell graph, $CG$, using a nuclei segmentation map and a tissue graph, $TG$, constructed by clustering superpixels into larger tissue areas based on similarity. To model the hierarchy, they introduced an assignment matrix $S_{CG \to TG}$, such that $S_{CG \to TG}(i,j) = 1$ if a cellular node $i$ from the cell graph belongs to tissue node $j$ in the tissue graph.
\\ \\
    Wang et al. \cite{wang2021hierarchical} introduced hierarchy by applying separate message passing operations on both a cell graph and a patch graph. As patch-level features, cellular node representations pooled based on the cells located in the patch. were used. The authors combined hierarchy learning with pre-established hierarchy by also applying self-attention graph pooling on both the cell- as well as the patch-graph. 
\\ \\
    Sims et al. \cite{sims2022using} connected a cell graph with a level-1 and level-2 patch graph, which represent patches of increasing size (400 $\mu$m, 800 $\mu$m). They define their message passing for any cellular node $i$ as $CG_i \longrightarrow L1_i \longrightarrow L2_i \longrightarrow L1_i \longrightarrow CG_i$, where each $\longrightarrow$ defines a message-passing function, $CG_i$ represents the node in the cell graph and $L1_i$,$L2_i$ represent the node corresponding to the level-1 patch and the level-2 patch on which this cell exists. By applying message-passing in this way, the model can exchange information between distant cells without using many message-passing layers, as the cellular nodes belonging to the same layer-2 node can be 800 $\mu$m away.
\\ \\
    Guan et al. \cite{guan2022node} proposed a \textit{Node-aligned} hierarchical graph-to-local clustering approach, inspired by the Bag-Of-Visual-Words (BOVW) methodology in Computer Vision. Starting with a set of H\&E stained WSIs, the authors first clustered the patches for each WSI, into visual word bags, where each bag is defined as $B$. A local clustering approach is used that samples global clusters from each bag $B$ into local subclusters using K-means. These subclusters represent a codebook of 'visual words' representing tissues with different properties. We can use this codebook to categorize patches in input WSIs into subclusters, from which we can construct a graph. This is achieved by connecting the patches in each subcluster using inner-sub-bag edges, and the subclusters themselves using outer-sub-bag edges. This graph structure allows hierarchically modeling WSIs by applying message-passing between patches in each subclusters to retrieve representations which are pooled on a subgraph-level. Subsequently, message-passing is performed between the pooled subcluster representations themselves.
\\ \\
    Hou et al. \cite{hou2022spatial} proposed constructing a cell graph along with superpixel-based tissue graphs at two levels ($CG, TG_{l1}, TG_{l2}$). They generated features for the cell graph by using a pretrained ResNet on a patch around the nucleus centroid, while generating tissue graph representations by averaging ResNet-embeddings from all crops belonging to a superpixel. The hierarchical information flow is modeled using a Transformer block that calculates the cross-attention between the graphs at different levels.
\\ \\
    Shi et al. \cite{shi2023structure} used graphs at four different levels of hierarchy: a tissue graph on 5x resolution, consisting of superpixels constructed using the SLIC algorithm \cite{achanta2012slic}, and 3 patch graphs at 5x, 10x, and 20x resolution, respectively. The 5x resolution patch graph is used to generate features for the tissue graph. Then, after applying message-passing to the 10x- and 20x patch graphs, the interaction between the different hierarchical levels is modeled using a hierarchical attention module. This module produces a tissue graph where the interactions are captured in the node features. Message-passing layers, global attention layers and a fully connected layer are applied subsequently to the tissue graph to come to a final prediction.
\\ \\
    Gupta et al. \cite{gupta2023heterogeneous} modeled a tissue graph and a cell graph together as a heterogeneous graph with cellular nodes, tissue nodes, cell-cell edges, tissue-tissue edges, and cell-tissue edges: $H = \{C,T,E_{cell \to cell}, E_{tissue \to tissue}, E_{cell \to tissue}\}$. After applying message-passing layers, they calculated the cross-attention between the cellular and tissue nodes using the transformer architecture to model the hierarchical relationships.  
\\ \\
    Abbas et al. \cite{abbas2023multi} established four separate hierarchical levels, where one level is a global image analyzed using a CNN model and the other levels are cell graphs constructed at different levels of semantic hierarchy (global, spanning the entire wsi ($G^{(0)}$), using patches of size $512x512$px ($G^{(1)}$) or using patches of size $256x256$px ($G^{(2)}$)). For each level, a subset of the segmented cells is randomly selected to build a cell graph. After applying message-passing layers on each level separately, the outputs are combined and processed using a fully connected layer. The combined representation and the representations gathered at each cell graph level separately are combined using an entropy weighting strategy, which weights the different representations based on the uncertainty of the model prediction given that representation. 
\\ \\

    \textbf{Multiresolution Hierarchies: }Xing et al. \cite{xing2021multi} constructed hierarchical patch graphs at several levels of image resolution, thus aggregating information from multiple resolution levels. Starting with a single patch, they subsampled the same patch at increasingly lower resolution, and connected the lower-resolution patches to the corresponding higher-resolution patch it was sampled from. This input graph was then used for a GNN model.
\\ \\
    Bazargani et al. \cite{bazargani2022multi} introduced hierarchy into their approach by constructing separate patch graphs on 5x, 10x and 20x resolution and then performing message-passing operations both on each graph separately as well as between graphs with different resolutions.
\\ \\
    Bontempo et al. \cite{bontempo2023mil} used a knowledge distillation approach combined with two patch graphs at different resolutions (high, low). They performed message-passing both hierarchically between high and low resolution and in each resolution graph itself. They treated the high-resolution graph as a 'teacher' and the low-resolution graph as a 'student' network, between which they optimize the KL-divergence for the bag-level predictions at each resolution.
\\ \\
    Mirabadi et al. \cite{mirabadi2024grasp} proposed modeling the pyramidal multi-magnification structure in whole slide images as a multiresolution graph, where information on both the inner-magnification and the intra-magnification levels could be modeled. They extracted patches from three magnification levels (20x, 10x and 5x), such that the patches on the higher resolutions are spatially equivalent to center crops of the patches at the lower resolutions. A RAG-graph was constructed such that nodes on each level were connected to both their adjacent neighbors on the same resolution as well as the spatially corresponding lower- and higher-level patch nodes. This allowed information to be exchanged between resolutions during message passing.  After message passing, a mean pooling operation was applied on each resolution level, resulting in a 3 node graph. This three-node graph embedding is then used for the downstream classification task.
\\ \\
    \begin{figure}[H]
        \centering
        \includegraphics[scale=0.45]{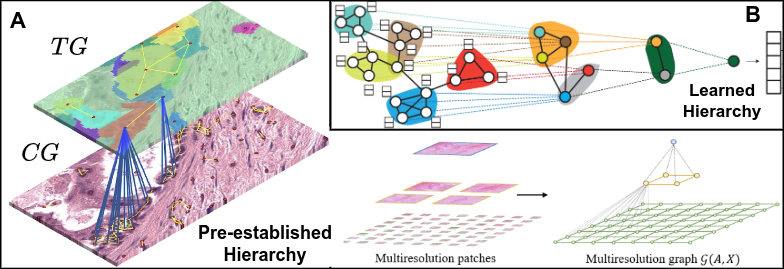}
        \caption{A) \textit{Pre-established hierarchy}, where different graphs are constructed at different levels of coarsening, which are connected hierarchically (e.g., using an assignment matrix) \cite{pati2020hact} \cite{xing2021multi}. B)\textit{Learned Hierarchy}, where trainable local pooling operations sequentially coarsen the graph structure \cite{ying2018graph}.}
        \label{fig:hierarchy}
    \end{figure}

       \begin{table}[H]
       \hspace{-3.5cm}
        \begin{tabular}{|c|c|c|c|}
        \hline
        \textbf{Publication} & \textbf{Date} & \textbf{Application} & \textbf{Hierarchy} \\
        \hline
        Pati et al. \cite{pati2020hact} & 2020/07 & ROI classification & $CG \to TG$ \\
        Xing et al. \cite{xing2021multi} & 2021/08 & Cancer subtyping & $PG_{40x} \to PG_{10x} \to PG_{5x}$\\
        Wang et al. \cite{wang2021hierarchical} & 2021/09 & Survival prediction & $CG \to PG$ \\
        Sims et al. \cite{sims2022using} & 2022/01 & ROI classification & $CG \to$ $PG_1 \to PG_2$ \\
        Hou et al. \cite{hou2022h} & 2022/06 & Binary classification & $PG_{10x} \to PG_{5x} \to PG_{thumbnail}$ \\
        Guan et al. \cite{guan2022node} & 2022/06 & Cancer subtyping & $S_k \to K_G \to B$ \\
        Hou et al. \cite{hou2022spatial} & 2022/09 & Cancer subtyping & $CG \to TG_{l1} \to TG_{l2}$ \\
        Shi et al. \cite{shi2023structure} & 2023/01 & Cancer grading & $PG_{20x} \to PG_{10x} \to TG_{5x}$ \\
        Wang et al. \cite{wang2023ccf} & 2023/02 & Cancer subtyping & $CG \to CCFG$ \\
        Gupta et al. \cite{gupta2023heterogeneous} & 2023/07 & Cancer subtyping, binary classification & $CG \to TG$ \\
        Bazargani et al. \cite{bazargani2022multi} & 2023/08 & Cancer subtyping & $PG_{20x} \to PG_{10x} \to PG_{5x}$ \\
        Bontempo et al. \cite{bontempo2023mil} & 2023/10 & Binary classification & $PG_{high} \to PG_{low}$\\
        Abbas et al. \cite{abbas2023multi} & 2023/12 & Cancer grading & $CG_{256px} \to CG_{512px} \to CG_{global} \to WSI_{thumbnail}$\\
        Mirabadi et al. \cite{mirabadi2024grasp} & 2024/02 & Cancer subtyping & $PG_{20x} \to PG_{10x} \to PG_{5x}$ \\
         \hline
        \end{tabular}
        \caption{Publications applying GNNs to histopathology which used a pre-established hierarchy. All hierarchies are shown small to large, such that when $X \to Y$, entities in $X$ are subordinate to the entities in $Y$. CG: Cell Graph, PG: Patch Graph, TG: Tissue Graph.}
        \label{tab:establishedhierarchies}
    \end{table}
   
   }
      %Overview of hierarchical graph representations and their relevance to histopathological analysis
      %Discussion on how hierarchical structures can enhance the understanding of tissue morphology

   \subsection{Adaptive Graph Structure Learning}
   \par{Most GNN applications in histopathology use a fixed input graph with fixed edge connectivity. While successful results have been achieved using this approach, we argue that it is suboptimal. Whether connections between nodes should exist is not clearly defined in the histopathology image, leading to the wide range of different approaches for constructing the input graphs, as previously discussed. These approaches are usually not based on biological or medical information, and thus introduce inductive bias which might not reflect the biology in the tissue. To counteract this problem, one can either adjust the message-passing equation such that some edges are given more representative power than others (e.g., using GAT \cite{velivckovic2017graph}), or one can make the graph construction a learnable transformation. The second approach, \textit{Adaptive Graph Structure Learning} (AGSL), has gained more popularity recently (Table \ref{tab:agsl}). In GNNs for histopathology, the AGSL strategy employs either a learned transformation that updates the adjacency matrix or learned convolutional filters that dynamically construct the graph.
   \\ \\
   \textbf{Learned Transformation:} In 2020, Adnan et al. \cite{adnan2020representation} introduced adaptive graph learning for the classification of lung cancer subtypes. The authors modeled the whole slide image as a fully connected graph of representative patches. Then, they used a pre-trained DenseNet for feature extraction. The graph connectivity is learned end-to-end using both global WSI context and local pairwise context between patches. Let us denote WSI $W$ with patches $w_1,...,w_n$, where for each patch $w_i$ we have a feature vector $x_i$. The authors first pooled the patch representations into a global context vector $c$ using a pooling function $\phi$ (e.g., sum):
   \begin{equation}
       c = \phi(x_1,x_2,...,x_n)
   \end{equation}
   The global vector $c$ is concatenated to each patch feature vector $x_i$ and is jointly processed by MLP layers which gives a feature vector $x_i^*$ that contains both local and global context information. Finally the matrix $X^*$, which holds all feature vectors $x^*$, is processed using a cross-correlation layer which determines the connectivity of the output graph in $A$, where each element $a_{ij} \in A$ represents the correlation between patches $w_i$ and $w_j$ and are used as edge weights in the learned graph structure. The learned graph can be used for any downstream tasks and has shown better performance than other (graph-based) MIL methods, available at the time.
   
   Hou et al. \cite{hou2022spatial} described a spatial-hierarchical GNN framework that could dynamically learn the graph structure during model training. Their \textit{Dynamic Structure Learning} module first embeds the representation of both node features $V$ and nuclear centroid coordinates $P$ together into a single representation $J$, using the following equation:
   \begin{equation}
       J = Concat[\sigma(P^T W_1), \sigma (V^T W_2)]
   \end{equation}
   Where $W_1$ and $W_2$ are learned weight matrices and $\sigma$ denotes a non-linear activation function. Next, the authors applied a distance-thresholded k-NN algorithm on the acquired embedding $J$ to determine the edge connectivity. Given a set of nodes $V$, set of edges $E$, distance threshold $d_{\text{min}}$ and the number of neighbors $k$, we use the following equation to determine the edges in $E$:
   \begin{equation}
        e_{uv} \in E \iff \{ u, v \in V \mid ||v - u||_2 \leq \min(d_k, d_{\text{min}}) \}
   \end{equation}
   Here, $d_k$ denotes the distance between nodes $u$ and the $k$-closest neighbor. 
\\ \\
    Liu et al. \cite{liu2023graphlsurv} propose learning the graph structure based on the cosine similarity between the transformed patch feature vectors. Given an input feature matrix $X$ and a transformation matrix $T$ we create a projected matrix $P = XT$. They then calculate the cosine similarity between each pair of patches in $P$ which are saved as a symmetric adjacency matrix $A_L$, which holds the 'edge strength' between any two patches in $P$. The edge strength is then thresholded using a set threshold $\epsilon$:
    \begin{equation}
e_{uv} \in E \iff \{ u, v \in V \mid \frac{{P[u] \cdot P[v]}}{{\|P[u]\| \cdot \|P[v]\|}} \leq \epsilon \}
    \end{equation}
    where $P[u]$ and $P[v]$ denote the projected feature vectors of nodes $u$ and $v$, respectively. Note that the transformation matrices are learned, which allows the graph structure to be adapted during model training. 
\\ \\
   \textbf{CNN-filter Based: }Gao et al. \cite{gao2022convolutional} and Ding et al. \cite{ding2023fractal} both use a very different approach, where the learned feature maps generated by a CNN are used as basis for the graph construction. More specifically, they treat the units in each feature map as nodes in which the features are spatially concatenated across channels into a node feature vector. After this concatenation, the K-nn algorithm is used to connect the nodes. By basing the graph structure on learned CNN feature maps, the graph structure is learned by training the CNN and, since each unit in the feature maps corresponds to a spatial region in the input image, the constructed graph can capture spatial dependencies between regions in the WSI. Given the acquired node embedding matrix $X \in \mathbf{R}^{N \times C}$ where $N$ is the number of nodes and $C$ the amount of channels, we determine the existence of edges as follows:
   \begin{equation}
       e_{uv} \in E \iff \{ u, v \in V \mid ||u_f - v_f||_2 \leq d_k \}
    \end{equation}
    Where $u_f$, $v_f$ are the feature vectors of node $u$ and $v$, and $d_k$ is the distance between node $u$ and the $k$-closest neighbor of $u$.
\\ \\
    %Explanation of dynamic graph structure learning techniques and their implications in histopathology

    \begin{table}[H]
        \hspace{-1.5cm}
        \begin{tabular}{|c|c|c|c|}
        \hline
        \textbf{Publication} & \textbf{Date} & \textbf{Application} & \textbf{Adaptive learning mechanism} \\
        \hline
        Adnan et al. \cite{adnan2020representation} & 2020/05 & Binary classification & Learned transformation \\
        Gao et al. \cite{gao2022convolutional} & 2022/02 & Cancer subtyping & CNN-filter based \\
        Hou et al. \cite{hou2022spatial} & 2022/09 & Cancer subtyping & Learned transformation \\
        Ding et al. \cite{ding2023fractal} & 2023/02 & Cancer subtyping, Cancer grading & CNN-filter based \\
        Liu et al. \cite{liu2023graphlsurv} & 2023/04 & Survival prediction & Learned transformation \\
        \hline
        \end{tabular}
        \caption{Publications applying GNNs in histopathology and using adaptive graph structure learning strategies.}
        \label{tab:agsl}
    \end{table}
   \subsection{Multimodal GNNs}
   \par{In histopathology diagnostics, different modalities are often combined to assist in clinical decision-making and prognostic predictions. While most applications of GNNs in histopathology focus solely on H\&E image data, approaches considering multiple modalities have gained popularity recently. Combining data from multiple modalities helps increase model accuracy and generalization. Graph Neural Networks are especially suitable for multimodal integration, as data from different modalities can be easily combined in the node- and edge feature vectors \cite{ding2022graph}. In the last few years, multiple approaches combined IHC-stained biopsy images with H$\&$E stained biopsy images, while other approaches incorporated spatial transcriptomics or genetic data in the model input. We differentiate between \textit{Stain multimodality}, where the same whole slide images with different stainings (e.g., IHC) are combined, and \textit{Full multimodality}, where the modalities are not based on WSIs (e.g., CT-scans, gene expression data). An overview of the multimodal GNNs in histopathology is given in Table \ref{tab:multimodal}.}

    %Fusion problem
   \par{An important challenge in multimodal integration in Deep Learning models is how- and where in the model architecture data from different modalities should be combined, which we call \textit{fusion}. In a GNN context, we broadly differentiate between \textit{early fusion}, where data from different modalities is combined prior to message passing and \textit{late fusion}, where data is combined after the message passing steps (Figure \ref{fig:modality_fusion}).
   
   We broadly categorize the multimodal GNNs into four groups: \textit{Pathomic fusion based}, which uses the pathomic fusion strategy, popularized by Chen et al. \cite{chen2020pathomic}, \textit{Early fusion}, \textit{Late fusion} and \textit{Modality prediction}, encompassing models that predict one modality using another. Models that do not directly fuse modalities but use predictions from one modality to drive how the other modalities are processed are considered Late fusion models.}

   \subsubsection{Full multimodality}
    \textbf{Pathomic Fusion:} Chen et al. \cite{chen2020pathomic} integrated whole slide image information together with RNA-Seq counts and copy number variant (CNV) information. They used this combined information for cancer subtyping and survival analysis on clear cell renal cell carcinoma and glioma TCGA datasets. Their multimodal model fused information from 3 different modules: A CNN-based image module, a GNN-based cell graph module, and a genomic module, which took CNV and RNA-seq information as input. In the image module, a set of WSI patches was used as input for an ImageNet-pretrained VGG19 CNN model optimized for cancer grading and survival prediction. The cell graph module first segmented the nuclei in the image, constructed a graph using these nuclei, and used message-passing layers to learn a graph representation. Lastly, the genomic module, where a self-normalizing neural network was learned on a vector of CNV- and RNA-seq information to learn a genomic representation. Their approach for multimodal fusion, which they call \textit{Pathomic fusion} models interactions between modalities via the Kronecker product of attention-gated representation. The attention gating is applied to the hidden representation of modality $m$, $h_m$, by learning a transformation $W_{ign \to m}$ which assigns an importance score for each modality, which we denote as $z_m$:
\begin{equation}
    \begin{aligned}
        h_{m,\text{gated}} &= z_m \ast h_m, \quad \forall m \in \{i, g, n\} \\
        \text{where,} \quad h_m &= \text{ReLU}(W_m \cdot h_m) \\
        z_m &= \sigma(W_{\text{ign}\rightarrow m} \cdot [h_i, h_g, h_n])
    \end{aligned}
\end{equation}
      Where $h_i$, $h_g$, and $h_n$, are the gated representation vectors of the image module, graph module, and genomic module, respectively. The authors calculated the Kronecker product of these vectors to get a combined representation $h_{fusion}$:
      \begin{equation}
          h_{fusion} =
\begin{pmatrix}
h_i \\
1 \\
\end{pmatrix}
\otimes
\begin{pmatrix}
h_g \\
1 \\
\end{pmatrix}
\otimes
\begin{pmatrix}
h_n \\
1 \\
\end{pmatrix}
      \end{equation}
    where $\otimes$ denotes the outer product. The result, $h_{fusion}$ is a three-dimensional tensor that can then be connected to a fully connected layer for classification tasks or survival prediction.
\\ \\
    Jiang et al. \cite{jiang2023predicting} predicted EGFR gene mutations in lung cancer by augmenting the approach used by Chen et al. \cite{chen2020pathomic}. The authors approach differs from Chen et al. by not using genomic data but instead using clinical information (e.g., gender, age) as the third modality, next to a spatial cell graph and whole slide image. Comparing with a previous model from the same group \cite{xiao2022lad}, which used a cell graph- and image module but no clinical features, the authors found considerable performance increases for the multimodal model.
\\ \\
    \textbf{Early Fusion: } Azher et al. \cite{azher2023spatial} integrated spatial transcriptomics data with accompanying H\&E WSI data to predict survival and grade cancer in colorectal cancer. The authors first constructed an embedding model that used an ImageNet-pretrained CNN to encode H\&E patches and fully connected layers to encode spatial gene expression data at the same location. They then optimized a projection layer to merge the data from these modalities into a single vector using a combination of unimodal and cross-modal SimCLR loss functions. This effectively trained the model to encode a cross-modal embedding vector. The acquired embeddings were used as node vectors in a GNN model for downstream tasks. The authors showed that using expression-aware embeddings improved model performance on all tasks, indicating that pretraining using coupled H\&E WSIs and spatial transcriptomics datasets can help retrieve more discriminative embeddings for downstream tasks.
\\ \\
    \textbf{Late Fusion:} Zuo et al. \cite{zuo2022identify} integrated H\&E stained WSIs with genomic biomarker information. Specifically, they constructed a graph of patches containing Tumor Infiltrating Lymphocyte (TILs) and analyzed this graph using a GNN. Genomic data consisted of mRNA gene counts, which were transformed to a gene co-expression module matrix using the lmQCM algorithm. They then applied a concrete autoencoder model to the co-expression matrix to identify survival-associated features. The GNN- and autoencoder outputs were then fused using a self-attention layer.
\\ \\
    De et al. \cite{de2022brain} combined MRI- and H\&E stained WSIs of brain tumors to predict the type of brain cancer. The modalities were not directly fused; instead, the authors first used a 3D-CNN model to detect whether the cancer was one of the possible cancer types (Glioblastoma). If this was the case, the model simply outputs glioblastoma as its prediction. When this was not the case, a patch graph was constructed from the H\&E image which was used as input for a GNN model. Finally, this GNN model could predict one of the remaining subtypes (Normal, Astrocytoma, or Oligodendroglioma).
\\ \\
    Xie et al. \cite{xie2022survival} combined gene expression with H\&E whole slide image data for survival prediction in gastric cancer. Here, the authors first processed ResNet-based WSI tile features and a gene expression matrix separately using MLP layers. Then the interaction between each WSI patch and each gene feature vector was calculated using a cross-modal attention layer. After this processing, the data from both modalities was aggregated using a MIL-aggregation module and finally fused using concatenation. The fused embeddings were used to construct a patient graph, based on the similarity of the fused embeddings between the patients. A GNN was used to process this graph, which produced a survival prediction.
\\ \\
    Zheng et al. \cite{zheng2023graph} fused gene-expression signatures with a WSI-patch graph using their \textit{Genomic Attention Module} approach. After message-passing on the patch graph, the pairwise interactions between each patch and each individual gene signature modeled using a self-attention mechanism. This allows the model to learn the interactions between spatial tissue regions and gene signatures, which allowed the authors to visualize which gene signatures were associated with certain regions in the WSI.
\\ \\
    \textbf{Modality Prediction:} Fatemi et al. \cite{fatemi2023inferring} integrated spatial transcriptomic data with co-localized H\&E WSI data to characterize spatial tumor heterogeneity in colorectal cancer. They achieved this by training a model to predict the spatial gene expression from the H\&E WSI. The authors tried to predict the spatial gene expression using both a CNN- and GNN-based network and showed that for this task, the CNN-based methods worked better.
\\ \\
    Gao et al. \cite{gao2023predicting} predicted spatial transcriptomic data using H\&E images by integrating image- and cell graph data using CNN- and GNN-based models. The authors showed that integrating the graph- and image-based information together did significantly improve over using either one alone.
\\ \\

    %Introduction to multimodal graph modeling and its significance in integrating diverse data types in histopathology
    %Examples demonstrating the integration of different modalities (e.g., images, text, graphs) for comprehensive analysis
    
    \subsubsection{Stain multimodality}
   \par{
    \textbf{Early fusion: }Li et al. \cite{li2022differentiation} fused information from Second-Harmonic Generation (SHG) microscopy images and H\&E WSIs together to differentiate between pancreatic ductal adenocarcinoma and chronic pancreatitis in pancreatic cancer. The images from both modalilities were registered and for each modality a separate graph was constructed. The features from each modality were combined into node features for the input graph, where nodes represented registered patches in both modalities. An ImageNet-pretrained ResNet model was used to retrieve features from the H\&E patches, while collagen fiber-specific handcrafted features were extracted for each SHG-patch. A H\&E-SHG graph was constructed where the node vectors contained the concatenation of the patch features from both modalities. This graph was used in a GNN model which predicted between the two classes.
\\ \\
   Gallagher-Syed et al. \cite{gallagher2023multi} integrated data from IHC- (CD138, CD68, CD20) and H\&E stained synovial biopsy samples to predict a Rheumatoid Arthritis subtype using a GNN model. Information between the staining modalities was exchanged by modeling each patch, from each staining, as a node and connecting the nodes based on their feature similarity to get a single multistain graph. The authors showed that the features across stains were similar enough to cause nodes from different staining to mix in the graph and, thus, enable information exchange between the modalities in message passing layers of the GNN. The authors used the multimodal graph as input for a GNN model whose output was used to predict the rheuma subtype.
\\ \\
   \textbf{Late fusion: }Dwivedi et al. \cite{dwivedi2022multi} combined trichrome- (TC) and H\&E stainings of liver biopsies to predict liver fibrosis.  The authors experimented with different modality fusion techniques. Their experiments showed that their late concatenation or addition and the pathomic fusion strategy proposed by Chen et al. \cite{chen2020pathomic} performed the best for fibrosis prediction. In the late and pathomic fusion strategies, they separately processed both the H\&E and TC tissues as graphs using a GNN and then fused the features from both modalities together.
   \\ \\
   Qiu et al. \cite{qiu2022intratumor} combined information from  H\&E stainings, multiphoton microscopy (MP), and two-photon excited fluorescence (TPEF) applied to the same breast cancer biopsies. Instead of fusing the modalities in the model itself, the authors determined tumor-associated collagen signatures from the 3 different modalities in different regions to calculate a 8-bit binary vector for each region. The regions sampled were treated as graph nodes having the binary vector as node attributes. Using these nodes, a fully connected graph was constructed and used as input for a GNN-model. The models output could be used for survival prediction.
   \\ \\

   \textbf{Modality prediction:} Pati et al. \cite{pati2023multiplexed} used a generative approach to virtually predict IHC stained tissue images from H\&E WSIs, and then used a multimodal GNN Transformer model to perform survival prediction and cancer grading tasks in prostate cancer, breast cancer, and colorectal cancer. The authors used three strategies for fusion (no fusion, early fusion, late fusion) and found that early fusion works optimally for both tasks. In early fusion, the authors combined ImageNet-pretrained ResNet features from the same patch in all modalities to form the node features in the input graph. In late fusion, meanwhile, all modalities were assigned a separate input graph, which was processed separately using the GNN Transformer model. Subsequently, the output features were combined. The authors hypothesized that early fusion allowed the model to learn multimodal spatial interactions during message passing, causing a performance gain compared to the other fusion strategy.
   }

   \begin{table}[H]
       \hspace{-4.4cm}
       \begin{tabular}{|c|c|c|c|c|}
       \hline
        \textbf{Publication} & \textbf{Date} & \textbf{Application} & \textbf{Fusion} & \textbf{Modalities}  \\
        \hline
        Chen et al. \cite{chen2020pathomic} & 2020/09 & Survival prediction, Cancer subtyping & Late (Pathomic fusion) & H\&E WSI, Gene expression, CNV \\
        Dwivedi et al. \cite{dwivedi2022multi} & 2022/04 & Cancer grading & Late & H\&E WSI, TC WSI\\
        Qiu et al. \cite{qiu2022intratumor} & 2022/07 & Survival prediction & Early & H\&E WSI, MP, TPEF \\
        Zuo et al. \cite{zuo2022identify} & 2022/09 & Survival prediction & Late (Self-attention) & H\&E WSI, Gene expression \\
        De et al. \cite{de2022brain} & 2022/10 & Cancer subtyping & None & H\&E WSI, MRI \\
        Li et al. \cite{li2022differentiation} & 2022/11 & Cancer subtyping & Early & H\&E WSI, SHG \\
        Xie et al. \cite{xie2022survival} & 2022/12 & Survival prediction & Late & H\&E WSI, Gene expression \\
        Fatemi et al. \cite{fatemi2023inferring} & 2023/03 & ST-prediction & None & H\&E WSI, ST \\
        Jiang et al. \cite{jiang2023predicting} & 2023/03 & Mutation prediction & Late (Pathomic fusion) & H\&E WSI, clinical data \\
        Gao et al. \cite{gao2023predicting} & 2023/07 & ST-prediction, survival prediction & None &  H\&E WSI, ST \\ 
        Gallagher et al. \cite{gallagher2023multi} & 2023/09 & Rheumatoid Subtyping & Early & H\&E WSI, IHC WSI\\
        Pati et al. \cite{pati2023multiplexed} & 2023/12 & Survival prediction, Cancer grading & Early, Late & H\&E, virtual IHC \\
        Azher et al. \cite{azher2023spatial} & 2024/01 & Survival prediction, Cancer grading & Early & H\&E WSI, ST \\
        Zheng et al. \cite{zheng2023graph} & 2024/01 & Survival prediction & Late & WSI, Gene Expression \\
        \hline
       \end{tabular}
       \caption{Applications of Multimodal GNNs in histopathology. CNV: Copy Number Variation, TC: Trichrome, MP: MultiPhoton microscopy, TPEF: two-photon excited fluorescence microscopy, MRI: Magnetic Resonance Imaging, SHG: Second-Harmonic Generation microscopy, ST: Spatial Transcriptomics, IHC: Immunohistochemistry}
       \label{tab:multimodal}
   \end{table}

\begin{figure}[H]
    \centering
    \includegraphics[scale=0.4]{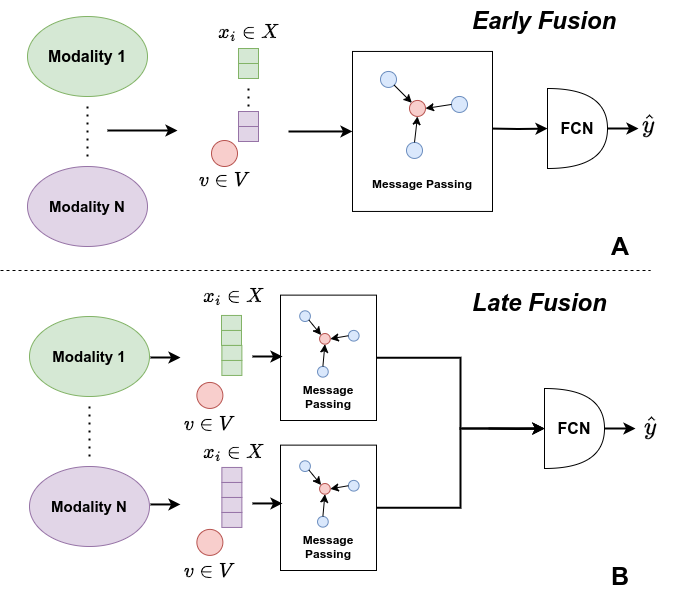}
    \caption{Early fusion (A) versus late fusion (B). In early fusion, information from different modalities is typically integrated in the node features before message passing, enabling modeling multimodal interactions. In late fusion, modalities are separately processed and combined before the final model layers which calculate the model prediction. FCN: Fully Connected Layer.}
    \label{fig:modality_fusion}
\end{figure}

\subsection{Higher-order graphs}
\par{While graphs have shown to be adequate formats for the representation of histopathology slides, it is limited by the fact only pairwise relations can be modeled. Furthermore, the entities in the graphs can solely be modeled as nodes and edges. This limitation has inspired extensions to the graph modeling framework, which are collectively known as higher-order graphs. Examples of higher-order graphs are hypergraphs, cellular complexes, and combinatorial complexes. To allow learning from these higher-order graph structures, message-passing frameworks called \textit{topological neural networks} (TNNs) have been developed \cite{papillon2023architectures}.}

\par{In histopathology, TNNs have not yet been widely adopted, but there has been a steadily increasing number of publications that model WSIs as hypergraphs. Hypergraphs extend the graph modeling framework with \textit{hyperedges}, which can connect sets containing an arbitrary number of nodes in the graph. This allows hypergraphs to model relations that rely on more than 2 pairwise entities. Deep learning on hypergraphs can be done using hypergraph neural network architectures, such as HGNN \cite{feng2019hypergraph} and HyperGAT \cite{ding2020more}. We provide an overview of publications using higher-order graphs in histopathology in Table \ref{tab:hypergraphpubs}.}

Let us denote a hypergraph as \( G = (V, E_{\text{hyp}}) \), which consists of a set of nodes \( V \) and a set of hyperedges \( E_{\text{hyp}} \). Each hyperedge in \( E_{\text{hyp}} \) is a pair of subsets of \( V \), allowing connections between any number of vertices. For example, a hypergraph with vertices \( V = \{v_1, v_2, v_3, v_4\} \) and hyperedges \( E_{\text{hyp}} = \{ \{v_1, v_2\}, \{v_2, v_3, v_4\}, \{v_1, v_3, v_4\} \} \) of \( V \), expressing relationships between multiple nodes simultaneously. We denote the connectivity of a hypergraph using an incidence matrix $H^{|V| \times |E|}$ whose entries are defined as:
\begin{equation}
    h(v,e) = \begin{cases}
        1, & \text{if } v \in e \\
        0, & \text{if } v \notin e \\
    \end{cases}
\end{equation}
for nodes $v \in V$ and edges $e \in E_{hyp}$. For any node $v$, its degree is defined as $d(v) = \sum_{e \in E_{hyp}} h(v,e)$, similarly for any edge $e \in E_{hyp}$, its degree is defined as $d(e) = \sum_{v \in V} h(v,e)$. These degrees are saved in diagonal matrices $D_e$ and $D_v$, which contain the edge degrees and node degrees, respectively. Lastly, we denote the matrix of node features as $X$. The decision on which nodes to connect to a hyperedge is usually based on the feature similarity or spatial distance (i.e. closely related nodes are connected together by a single hyperedge). Feng et al. \cite{feng2019hypergraph} introduced the hypergraph neural network (visualized in Figure \ref{fig:hgnn}), which defined a message passing operation on hypergraphs as follows:
\begin{equation}
 \mathbf {X}^{(l+1)}=\sigma \left( \mathbf {D}_{v}^{-1 / 2} \mathbf {H} \mathbf {W} \mathbf {D}_{e}^{-1} \mathbf {H}^{\top } \mathbf {D}_{v}^{-1 / 2} \mathbf {X}^{(l)} \mathbf {\Theta ^{(l)}}\right) 
\end{equation}
where $W$ is a learned weight matrix, $\sigma$ denotes a nonlinear activation function, and $\Theta$ is a learnable filter matrix used for feature extraction. After applying message passing, we have an updated feature matrix $X$. This can then be used to obtain features on the hyperedge level using the equation $X^{(l+1)}_{he} = H^T \times X$. Finally, the updated node-level embeddings are acquired by multiplying the hyperedge features with the incidence matrix: $X^{(l+1) \prime} = X^{(l+1)}_{he} \times H$.   

\begin{figure}[H]
    \centering
    \includegraphics[scale=0.4]{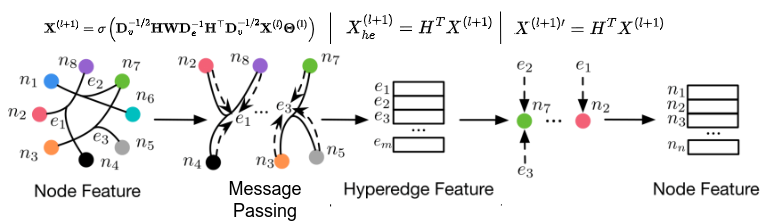}
    \caption{Graphical overview of the hypergraph neural network framework \cite{feng2019hypergraph}. First message-passing gets applied between all nodes connected to the same hyperedge. Then the learned features are calculated on an hyperedge-levels. Finally, the hyperedge-level features are used to calculate the new node features.}
    \label{fig:hgnn}
\end{figure}
\par{Di et al. \cite{di2020ranking} were the first to model WSIs as hypergraphs. They used their hypergraph approach for survival prediction in lung and brain cancer datasets. The authors started by constructing sets of $K$ similar patches based on the Euclidean distance between the feature vectors, which were retrieved using an ImageNet-pretrained ResNet model. $N$ hyperedges are then used to connect the patches in each of the sets. The authors then used the node feature matrix $X$ with the defined hypergraph, captured in $H$, and updated the features using a series of convolutional hypergraph layers (HGNN). The acquired representations after message-passing are then used for the downstream survival prediction task. The authors show that their hypergraph-based method outperforms other CNN- and GNN-based frameworks for survival prediction.}
\\
\par{Bakht et al. \cite{bakht2021colorectal} followed by the construction of a patch-based hypergraph for the classification of patches in colorectal cancer. They used an ImageNet-pretrained VGG-19 model for extracting features for each WSI patch. Given a fixed neighbor parameter $k$, their hypergraph construction strategy starts by defining the distance between any two patches $i$, $j$ as:
\begin{equation}
    d_k(i, j) = \exp\left(-\frac{||x_i - x_j||_2^2}{2\sigma^2}\right)
\end{equation}
where $x_i$, $x_j$ represent the feature vectors of patch $i$ and $j$, respectively, and $\sigma$ is a bandwidth parameter. Then, the authors calculated the vertex-edge probabilistic incidence matrix which determines the probability of a node $v$ to be connected using hyperedge $e$:
\begin{equation}
    h(n, e) = \begin{cases} 
  \exp\left(-\frac{d}{p_{\text{max}}d_{\text{avg}}}\right), & \text{if } v \in e \\
  0, & \text{if } v \notin e 
\end{cases}
\end{equation}
Here $d$ denotes the distance between the current node $n$ and the neighboring node. $p_{max}$ denotes the maximum probability and $d_{avg}$ is the average distance between all $k$ nearest neighbors. Finally, they use this incidence matrix to calculate the node and edge degrees:
\begin{equation}
d(v) = \sum_{v \in V} h(v, e), \quad d(e) = \sum_{e \in E} h(v, e)
\end{equation}
The degrees are combined into matrices $D_v$ and $D_e$, which are used, together with the incidence matrix $H$ and node feature matrix $X$ in 3 HGNN message passing layers. The output of these layers was used to predict the label of patches in the WSI.
  }
\\ \\
Di et al. \cite{di2022generating} then expanded on their previous work by using multiple hypergraphs that are fused together to be used as input for message passing layers. Specifically, they construct a topological hypergraph and a phenotype (feature-based) hypergraph. The authors sampled patches sequentially from the tissue boundary to the tissue center and grouped the patches in the same sequence step in the same topological area. The topological hypergraph is constructed by connecting neighboring patches with a hyperedge if they belong to the same topological area. The phenotype hypergraph meanwhile, is constructed using K-NN based on the vector similarity between the patch features. The two hypergraphs are then concatenated together to form a total incidence matrix $H$. For processing the constructed hypergraph, the authors use max-mask convolutional layers, which are defined in 4 iterative steps:
\begin{enumerate}
    \item \textbf{Hyperedge Feature Gathering} First, hyperedge-level features are formed by multiplying the hypergraph incidence matrix (\( H \)) and the node feature matrix (\( F_v^{(l)} \)). This step aggregates the information from nodes connected by each hyperedge, resulting in hyperedge-level features (\( F_e^{(l)} \)).
    \item \textbf{Max-Mask Operation} After gathering hyperedge-level features, a max-mask operation is performed on each dimensionality of \( F_e^{(l)} \). This operation aims to avoid overfitting by disregarding the contribution of dominant hyperedges that take the largest values.
    \item \textbf{Node Feature Aggregating} By multiplying the hyperedge features with the transposed incidence matrix ($H^T \times F_e^{(l)}$), we can calculate the node features (\( F_v^{(l+1)} \)).
    \item \textbf{Node Feature Reweighting} Finally, the output node features are further weighted using learnable parameters (\( \iota^{(l)} \)), which are represented as a diagonal matrix. This reweighting is followed by a non-linear activation function (\(\sigma\)). The reweighting step allows the model to learn the importance of different node features and adaptively adjust them.
\end{enumerate}

Mathematically, the max-mask convolutional layer is defined as follows:
\begin{equation}
\begin{aligned}
X^{(l+1)} &= \sigma \left( (I - L)X^{(l)} + H^{-1}(I - L)X^{(\lambda)} \cdot \iota^{(l)} \right) \\
F^{(l+1)}_e &= H^{-1}(I - L)X^{(l)} + X^{(\lambda)}
\end{aligned}
\end{equation}
Here, $L$ is the multigraph Laplacian matrix, and $I$ denotes the identity matrix. $H^{-1}(I - L)X^{(\lambda)}$ functionally ensures that the top $\lambda$ attribute feature dimensionalities are ignored during gradient calculation.
\\ \\
Bankirane et al. \cite{benkirane2022hyper} used adaptive agglomeration clustering to construct a patch hypergraph, which was then processed using a combination of HGNN and HGAT layers. The authors used self-supervised learning to learn patch-level representations. For agglomeration clustering, a similarity kernel was used that took into account both spatial locality and feature similarity between patches. This kernel calculated similarity scores between all two patches. If the similarity score was higher than a fixed threshold $\delta$, the patches were assigned to the same cluster $C_k$. For each cluster, the representation of the patches in the cluster was averaged to obtain cluster-level representations. Each clustered patch is treated as a node of a hypergraph. The hyperedges connected all nodes with a feature similarity higher than a fixed threshold $\delta_h$. We denote the neighborhood of a clustered node $c_i$ as $\gamma(c_i) = {c_i \in C ; \kappa_h(c_i,c_j) \geq \delta_h}$. Here, $C$ denotes the set of all clusters and$\kappa_h(c_i,c_j)$ denotes the output of the feature similarity kernel $\kappa_h$. Having determined the neighborhood, we can calculate the incidence matrix $H$ where:
\begin{equation}
    h_{k,j} = \begin{cases}
        1, & \text{if } c_i \in \gamma(c_i) \\
        0, & \text{else} \\
    \end{cases}
\end{equation}
The authors then used the incidence matrix $H$, and node feature matrix $X$ as input for a series of HGNN-HGAT layers and were finally pooled into a hypergraph-level representation. This representation was finally used as input for an MLP layer which predicted the hazard score for survival prediction.
\\ \\
Most recently, Liang et al. \cite{liang2024caf} introduced the \textit{adaptive HGNN} to histopathology, for the classification of sentinel node metastases and the differentiation between lung squamous cell carcinoma and lung adenocarcinoma. Here, the authors used the K-NN algorithm on patch-level ImageNet-pretrained ResNet features to construct a hypergraph of patches, where the $k$ most similar patches were connected using a hyperedge. Their main innovation comes in the form of adaptive HGNN, which can adjust the correlation strength between nodes and hyperedges on the graph during model training. They first denote a matrix of edge strength in layer $l$ as $T^{(l)}$. Each element $t_{i}^{(l)} \in T^{(l)}$, which denotes the attention score of the node $i$ and its associated hyperedge $e_{i,i^{\prime}}$ in the $l$-th layer, is defined as:
\begin{equation}
    t_{i}^{(l)} = \frac{{\exp (\sigma (sim(f_{i} M^{(l)} ,e_{{i,i^{\prime}}}  M^{(l)} )))}}{{\sum\nolimits_{{k \in N_{j} }} {\exp (\sigma (sim(f_{i} M^{(l)} ,e_{i,k} M^{(l)} )))} }}
\end{equation}
here, $M^{(l)}$ denotes a feature transformation matrix. $e_{i,i'}$ denotes the hyperedge in connecting node $i$ and $i^{\prime}$. By calculating these edge strength scores, the incidence matrix can be updated as follows:
\begin{equation}
\tilde{H}^{i{\prime\prime(l)}} = D_{V}^{ - 1/2} (T_{i}^{(l)} \odot H^{i} )WD_{e}^{ - 1} (T_{i}^{(l)} \odot H^{i} )^{{\text{T}}} D_{V}^{ - 1/2}
\end{equation}
where $D_v$, $D_e$ denote the node degree and edge degree matrices. $T^{(l)}$ denotes the edge strength matrix and $W$ is a learnable weight matrix. This function essentially adapts the node interconnection in $H$ using the calculated edge strengths in $T^{(l)}$. Note that the edge strength changes depending on the layer $l$, as the feature similarities also change between layer embeddings. The feature matrix is updated as follows:
\begin{equation}
    \tilde{F}_{i}^{(l + 1)} = \{ \tilde{f}_{i,j} \}_{j = 1}^{P} = \sigma ((\tilde{H}^{i{\prime\prime(l)}} )F_{i}^{(l)} P_{i}^{(l)} )
\end{equation}
where $\sigma$ is a nonlinear activation function and $P_i^(l)$ denotes a learned projection matrix. 
\begin{table}[H]
    \hspace{-1.5cm}
    \begin{tabular}{|c|c|c|c|c|c|}
    \hline
    \textbf{Publication} & \textbf{Date} & \textbf{Application} & \textbf{Hypergraph type} & \textbf{Message-Passing} \\
    \hline
    Di et al. \cite{di2020ranking} & 2020/09 & Survival prediction & Patch Hypergraph &  HGNN \\
    Bakht et al. \cite{bakht2021colorectal} & 2021/05 & Patch classification & Patch Hypergraph & HGNN \\
    Di et al. \cite{di2022generating}& 2022/09  & Survival prediction & Patch Hypergraph & HGMConv \\
    Benkirane et al. \cite{benkirane2022hyper} & 2022/11 & Survival prediction & Patch Hypergraph & HGCN, HGAT \\
    Liang et al. \cite{liang2024caf} & 2024/02 & Binary classification & Patch HyperGraph & Adaptive HGNN \\
    \hline
    \end{tabular}
    \caption{Publications which utilized hypergraph neural networks for histopathology WSI analysis.}
    \label{tab:hypergraphpubs}
\end{table}

\section{Future Prospects and Directions}
    \subsection{Topological Deep Learning}
    \par{In our review, we highlighted the application of deep learning on hypergraphs in histopathology. Interestingly, this approach has only been applied on a patch level, whereas we argue that hypergraph-based modeling might be very well suited for cell-level modeling. For example, cells can be organized in clusters that can have an important diagnostic context \cite{Chandran2012Cluster}. Such cell clusters could be modeled using hypergraphs, where homogeneous clusters are connected using a single hyperedge. Furthermore, there exist many other higher-order graph types such as cellular complexes and combinatorial complexes. We anticipate that these approaches will also be tested in a histopathological context. For example, using cellular complexes, different semantic tissue structures (e.g., tertiary lymphoid structures) can be modeled jointly with cells, but as separate graph entities.}
    %Few papers applied Hypergraphs, but other parts of TDL remain unused in histopathology
    %TDL can explicitly model hierarchies, how can this be used for histopathology?
    \subsection{Graph transformers}
    %Exphormers, MAMBA models, Do we even need graph structure if we can model all nodes as fully connected?
    \par{In the last few years, GNNs have been combined with transformer architectures, which has given birth to the Graph Transformer modeling paradigm. Graph transformers either use the positional embedding of the graph in the input to the transformer module, use the graph structure as a prior to build an attention mask for each input, or directly combine message passing layers with transformer blocks in the model architecture \cite{min2022transformer}. Graph transformers are especially suited for modeling long-distance relations in graphs, as they do not suffer from \textit{oversmoothing}, where node representations become almost identical across the graph when using increased GNN layer depth and \textit{oversquashing}, where the computational costs of adding GNN layers growth exponentially \cite{kreuzer2021rethinking}. In histopathology, these graph transformers have also been used. One major challenge in the application of graph transformers is their scalability, as the time- and memory complexity of the attention mechanism in Transformers grows exponentially ($O(|V|^2)$, where $V$ is the number of nodes). This is especially a problem in cell graphs in histopathology, as these graphs often pass 10.000 nodes in size. Recently, efforts have been made to greatly mitigate this scalability challenge \cite{rampavsek2022recipe} \cite{shirzad2023exphormer} \cite{wu2024simplifying}, which leads us to believe the popularity of graph transformers in histopathology will continue to grow.}
    \subsection{Graph-based multimodality}
    %Multimodal integration using GNNs, Patient Graphs
    In our review, we highlighted the use of graph-based modeling in multimodal approaches, but we argue that graphs themselves should be utilized more for the multimodal integration itself. For example, several researchers have used the concept of a \textit{Patient graph}, where nodes represents (aggregated) datapoints from different medical modalities corresponding to the same patient \cite{kim2023gnn} or multiple patients \cite{gao2020mgnn} \cite{ochoa2022graph}. Some approaches use graphs to model time series data, where, for example, medical information on the same patient gathered at different timepoints can be effectively utilized \cite{rocheteau2021predicting} \cite{daneshvar2022heterogeneous}.
    Zheng et al. proposed a framework in which adaptive graph structure learning and GNNs are combined to integrate data from different medical modalities for disease prediction \cite{zheng2022multi}. One major problem in the application of multimodal approaches in histopathology is that, often, not every modality is available for each patient. This effectively creates a missing modality problem. Ma et al. proposed a Bayesian meta-learning framework which mitigates this problem, allowing effectively multimodal learning and prediction even when a large number of modalities are missing in the data \cite{ma2021smil}. We argue that these approaches should be combined to effectively model the relationships between modalities, based on the task at hand, even in settings where modalities are missing from the patient data.
    
    \subsection{SSL using GNNs}
    %SSL is been used for feature extraction, but not many methods explored SSL on the graph itself.
    \par{Due to the high costs of annotation in histopathology, adaptation of self-supervised learning (SSL) has been steadily growing in histopathology applications, particularly for feature extraction. As such, they have been primarily adapted for feature extraction in GNN approaches. Although there have been a handful of approaches that used graph-based SSL \cite{ozen2021self} \cite{pina2022self}, we argue that more can still be gained from this approach. For example, only contrastive approaches have been tried, which leaves room for other schemes (e.g., autoregressive, generative). We propose using an approach similar to that of Deep Graph Infomax \cite{velivckovic2018deep}, where the aim is to maximize the mutual information between the global graph structure and the local subgraphs. This effectively makes the node features mindful of the global graph structure. This idea can be extended in hierarchical histopathology graphs, where the agreement between intermediate graphs (e.g., cell graphs and tissue graphs) can be maximized, to get more context-aware embeddings, similar to work by Yan et al. \cite{yan2023hierarchical}} 
    \subsection{Hierarchical modeling in GNNs}
    %Need for histopatholgoy hierarhchical GNN specific tools, which can view the graph at different coarsity levels and thus provide explainability on multiple levels.
    \par{As explained in our review, hierarchical GNNs are an increasingly popular modeling technique for histopathology WSIs, due to the information in WSIs existing on different levels of coarsity. We believe that this trend will continue and extend to hypergraphs and other higher-order graph structures, for which hierarchical pooling frameworks are currently being established \cite{zhao2023self} \cite{cinque2023pooling}. Another future approach will be to learn the necessary level of coarsening to establish an effective hierarchical structure end-to-end, which is currently controlled using a pooling ratio hyperparameter. We argue that different levels of graph coarsity might be optimal for different problems, as some problems in histopathology rely more on cell-level information, while others more on larger tissue structures. Lastly, in most current approaches, message-passing occurs on each level of hierarchy separately, not directly between hierarchies. We argue that the field could move to message passing schemes that are more effective at taking into account the hierarchical graph structure \cite{zhong2023hierarchical}.}
    \subsection{Foundation models in computational histopathology}
    \par{The rise of self-supervised learning as well as increased availability of histopathology datasets, has allowed the construction of very large deep neural networks, termed \textit{Foundation modes}, trained on huge amounts of (unlabeled) histopathology images \cite{vorontsov2023virchow}. These models can be used for effective feature extraction in a wide variety of tissue types. Recent approaches have introduced medical texts in addition to image data \cite{lu2023towards} \cite{huang2023visual}, which allows associating image data with medical texts and is thus very suitable for CBHIR applications. In both natural language processing and computer vision, there has been a move to foundation models that incorporate an even broader spectrum of modalities (video \cite{christensen2023multimodal}, audio \cite{gardner2023llark}, knowledge graphs \cite{luo2023molfm}). We argue that in histopathology and medical imaging in general, there will also be a move towards broader multimodality, especially given the amount of different modalities available in the medical domain (WSI, IHC, MRI, CT, EHR, etc.). Graph models of WSIs can also be used as input in these models, to encode the topological information present in WSIs and correlate that with the image data.
    %SSL trained foundation models and what their role could be in the future of computational histopathology
    %How do graphs fit into that picture? Multimodal integration?
    \subsection{Adaptive graph structure learning}
    \par{We have seen that adaptive graph structure learning is currently based either on learned projections or CNN filters. Outside of histopathology, most adaptive graph structure learning assume graph \textit{homophily} \cite{zhu2021survey} where similar nodes are likely to be colocated. In histopathology, this is not always the case, as some structures might be composed of different cell types which can vary widely in morphology. Furthermore, most applications focus on homogeneous graphs, where a single type of node and edge exists. Work by Zhao et al. \cite{zhao2021heterogeneous} showed that we can learn a heterogeneous graph optimized for downstream tasks, which is suitable for graphs showing heterophily, which can be the case in histopathology. Therefore, we argue that heterogeneous graph learning will be a useful approach for histopathology, if we model the WSIs as a heterogeneous graphs. 
}
    %How will the adaptive learning of graph structure evolve
    %Heterogeneous graphs, 

   %Potential applications of emerging trends in GNNs for histopathology
   %Suggestions for future research directions and areas for improvement
   %Importance of interdisciplinary collaborations in advancing GNN-based histopathological analysis

\section{Conclusion}
\par{In this review, we provided a comprehensive overview of the recent developments in the applications of GNNs in histopathology, which can be used for guiding new research in the field. We quantified the growth of different modeling paradigms in the use of GNNs in histopathology. Based on our quantification, we provided a comprehensive overview of several emerging subfields, including hierarchical graph models, adaptive graph structure learning, multimodality using GNNs, and higher-order graph models. We then provided future directions in the field, including the use of topological deep learning, graph transformer models, self-supervised learning using GNNs, the use of foundation models and expanding adaptive graph structure learning to heterogeneous graphs. }

\bibliography{bib}
\bibliographystyle{unsrt}
\end{document}